# Graph Signal Processing for Heterogeneous Change Detection–Part II: Spectral Domain Analysis

Yuli Sun, Lin Lei, Dongdong Guan, Gangyao Kuang, *Senior Member, IEEE*, Li Liu, *Senior Member, IEEE*

*Abstract*—This is the second part of the paper that provides a new strategy for the heterogeneous change detection (HCD) problem, that is, solving HCD from the perspective of graph signal processing (GSP). We construct a graph to represent the structure of each image, and treat each image as a graph signal defined on the graph. In this way, we can convert the HCD problem into a comparison of responses of signals on systems defined on the graphs. In the part I, the changes are measured by comparing the structure difference between the graphs from the vertex domain. In this part II, we analyze the GSP for HCD from the spectral domain. We first analyze the spectral properties of the different images on the same graph, and show that their spectra exhibit commonalities and dissimilarities. Specially, it is the change that leads to the dissimilarities of their spectra. Then, we propose a regression model for the HCD, which decomposes the source signal into the regressed signal and changed signal, and requires the regressed signal have the same spectral property as the target signal on the same graph. With the help of graph spectral analysis, the proposed regression model is flexible and scalable. Experiments conducted on seven real data sets show the effectiveness of the proposed method.

## I. INTRODUCTION

In the heterogeneous change detection (HCD), because the multi-temporal images are acquired by different sensors and show quite different appearances and characteristics, the core of HCD is to find the connection between heterogeneous images to make them comparable. As introduced in the part I [1], there are two main challenges in HCD under the unsupervised condition: first, how to find a robust and universal connection (transformation) between heterogeneous images, which is stable even if the HCD scene is very complex; second, how to alleviate the negative influence of the unknown changed samples in the transformation.

In the part I of this paper [1], we have introduced the idea of solving the HCD from the perspective of graph signal processing (GSP). We construct a graph for each image to capture the structure information, and then treat each image as the signal on the graph. The changes between images caused by the event produce two outcomes: first, changes in the graph;

Y. Sun, L. Lei and G. Kuang are with College of Electronic science, National University of Defense Technology, Changsha 410073, China.
D. Guan is with the HighTech Institute of Xi'an, Xi'an 710025, China.
L. Liu is with the College of System Engineering, National University of Defense Technology, Changsha, 410073, China.
This work was supported in part by the National Key Research and Development Program of China No. 2021YFB3100800 and the National Natural Science Foundation of China under Grant Nos. 61872379, 12171481 and 61971426.

second, changes in the graph signal. In this way, we can analyze the responses of different signals (*i.e.*, $\mathbf{X}$ and $\mathbf{Y}$) on different systems (*i.e.*, $H(\mathbf{S}_{t1})$ and $H(\mathbf{S}_{t2})$) defined on the graphs to detect the changes: 1) we compute the changes by measuring the structure difference between the graphs from the vertex domain (in part I), which is calculated by comparing the output signals of the same signal on different graph filters, *i.e.*, the differences between $H(\mathbf{S}_{t1})\mathbf{Y}$ and $H(\mathbf{S}_{t2})\mathbf{Y}$, or $H(\mathbf{S}_{t1})\mathbf{X}$ and $H(\mathbf{S}_{t2})\mathbf{X}$; 2) we can compute the changes in graph signals between $\mathbf{X}$ and $\mathbf{Y}$ by comparing the spectral properties of different signals on the same graph filter, *i.e.*, the difference between $H(\mathbf{S}_{t1})\mathbf{X}$ and $H(\mathbf{S}_{t1})\mathbf{Y}$, or $H(\mathbf{S}_{t2})\mathbf{X}$ and $H(\mathbf{S}_{t2})\mathbf{Y}$ in the spectral domain (in this part II), as shown in Fig. 1.

To avoid the leakage of heterogeneous data, in the measuring of structure difference of part I [1], we weaken the effect of difference in the original pixel values on image structure, and focus more on the changes in the connectivity between vertices. Therefore, we choose the graph shift operator $\mathbf{S}$ as the normalized average weighting matrix $\mathbf{W}^{\mathrm{avg}}$, or normalized random walk matrix $\mathbf{P}$ and Laplacian matrix $\mathbf{L}^{\mathrm{rw}}$ to measure the change level, for example, $\mathbf{d}_i^{\mathbf{x}} = (\mathbf{W}_{t2}^{\mathrm{avg}}\mathbf{X})_i - (\mathbf{W}_{t1}^{\mathrm{avg}}\mathbf{X})_i$. Similarly, directly comparing $H(\mathbf{S}_{t1})\mathbf{X}$ and $H(\mathbf{S}_{t1})\mathbf{Y}$ will also cause the leakage of heterogeneous data. However, Setting $\mathbf{S} = \mathbf{W}^{\mathrm{avg}}$ at this time does not solve the problem. Alternatively, we analyze the graph signal on the spectral domain, which can eliminate the influence of heterogeneous data from different domains.

Nevertheless, spectral domain analysis also presents another challenge: how to get the change of signal from the difference of spectral domain. Since the goal of HCD is to find the region where the change occurred during the event, it is directly corresponding to the region represented by the vertex in the graph. Therefore, the change measurement of $\mathbf{d}_i^{\mathbf{x}} = (\mathbf{W}_{t2}^{\mathrm{avg}}\mathbf{X})_i - (\mathbf{W}_{t1}^{\mathrm{avg}}\mathbf{X})_i$ in part I that finds the changes in the vertex domain can directly output the HCD results: the changed region. As we known, in the classical signal processing, the Fourier coeffcients are the integration or summation form of the signal in the time domain. Similarly, the graph frequency coeffcients are also the summation form of the graph signal in the vertex domain. Therefore, directly comparing the spectra of two graph signals in the spectral domain is not possible to find the changed vertex (*i.e.*, changed region).

To address the challenge introduced by the spectral analysis, we use a signal decomposition method: we decompose the source signal $\mathbf{Y}$ into the regressed signal $\mathbf{Z}$ and the changed signal $\mathbf{\Delta}$ as $\mathbf{Y} = \mathbf{Z} + \mathbf{\Delta}$, and require the regressed signal

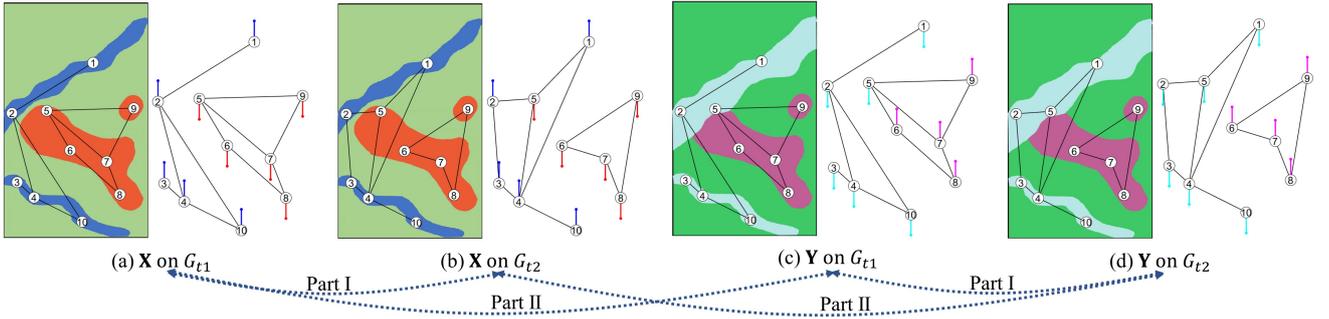

(a) $\mathbf{X}$ on $G_{t1}$    (b) $\mathbf{X}$ on $G_{t2}$    (c) $\mathbf{Y}$ on $G_{t1}$    (d) $\mathbf{Y}$ on $G_{t2}$

Fig. 1. Two strategies for the HCD from GSP perspective: calculating the structure difference by letting the same graph signal pass through different graph filters (part I); and calculating the signal difference by letting different graph signals pass through the same graph filter (part II).

$\mathbf{Z}$ have the same spectral property as the target signal $\mathbf{X}$ on the graph $G_{t1}$. With this decomposition model, we can obtain the changed signal $\mathbf{\Delta}$, which can be directly used to detect the areas of change. On the other hand, if we treat $\hat{\mathbf{Z}}$ as the regression image, because the regressed signal $\mathbf{Z}$ is separated from $\mathbf{Y}$, then $\hat{\mathbf{Z}}$ and $\hat{\mathbf{Y}}$ belong to the same domain. And because $\mathbf{Z}$ and $\mathbf{X}$ have the same spectral property, their structure is consistent. Therefore, this signal decomposition model can also be regarded as an image regression method.

### A. Related image regression works

It is a natural and intuitive idea to use an image regression approach to solve HCD problems, which transforms one image to the domain of the other image, and thereby converts the HCD into homogeneous change detection (CD) by comparing the regressed image and the target domain image. The homogeneous pixel transformation (HPT) method builds up the mappings between pixel values of heterogeneous images with a K-nearest neighbors (KNN) based kernel regression [2], which is supervised by the unchanged samples. To avoid the reliance on labeled data, an affinity matrix difference (AMD) based image regression method (AMD-IR) is proposed [3], which first employs the AMD to pick probably unchanged pixels as the pseudo-training set, and then uses different regression methods to complete the image translation. Some deep learning based regression methods have also been proposed for HCD, such as the coupling translation network (CPTN) [4], deep translation based CD network (DTCDN) [5], conditional generative adversarial network (cGAN) [6], AMD based X-Net and adversarial cyclic encoder network (ACE-Net) [7], cycle-consistent adversarial networks (CycleGANs) based method [8], and the image style transfer-based method (IST) [9].

The proposed signal decomposition based regression method has three attractive features when comparing with the above regression-based HCD methods: first, different from the common regression-based methods that aim to learn a luminance transformation to map one image to the domain of the other image, i.e., the pixel-to-pixel transformation, the proposed method is a structure regression based on the spectral analysis, which is robust to the noise and complex HCD conditions. Second, the unknown changed samples will affect the unsupervised regression process, whether manually constructed or learned regression functions. Since it is impossible to know in advance which pixels are changed, these methods can only train the regression function by selecting a pseudo-training set (e.g., AMD-IR [3], X-Net and ACE-Net [7]) or by treating all samples (pixels) as unchanged and then correcting the regression function step by step with an iterative process to refine the training samples (e.g., cGAN [6] and IST [9]), which is a coarse-to-fine process. However, the signal decomposition model, $\mathbf{Y} = \mathbf{Z} + \mathbf{\Delta}$, directly outputs both the regression signal $\mathbf{Z}$ and changed signal $\mathbf{\Delta}$, which means that the changed samples are removed from the original $\mathbf{Y}$, so the negative influence of the changed samples on the regression process is reduced. Third, the proposed signal decomposition based regression method is very intuitive and interpretable by using the graph spectral analysis, and it is also very flexible by designing graph filters with different spectral characteristics.

### B. Contributions

The main contributions of this part II are as follows.

- We analyze the spectral properties of different images on the defined graphs, and illustrate the connection between changes in heterogeneous images and differences in their spectral properties.
- We propose a signal decomposition based regression method for HCD based on the spectral analysis, which decomposes the source signal into the regressed signal and changed signal, and constrains the spectral properties of the regressed signal and target signal to be consistent.
- We analyze the vertex domain filtering (VDF) based HCD method proposed in part I in the spectral domain. We also give some discussions about the proposed regression method, and show that the proposed method can also be extended to other applications.
- Experimental results on seven data sets demonstrate the effectiveness of the proposed method by comparing with some state-of-the-art (SOTA) methods (source code will be made available at https://github.com/yulisun/HCD-GSPpartII).

### C. Outline

The remainder of this part is structured as follows: Section II describes the related basics of GSP. Section III analyzes the spectral property of the heterogeneous images on different graphs. Section IV proposes the signal decomposition based regression method based on the spectral domain analysis. Section V gives some discussions of the proposed method. Section VI presents the experimental results. Finally, Section VII concludes this paper and mentions the future work. For convenience, Table I lists some important notations used in the rest of this part. One notation should be noted: $\mathbf{X}_i$ represents the $i$-th **row** of matrix $\mathbf{X}$ in this paper, not the $i$-th column as commonly used.

TABLE I
LIST OF IMPORTANT NOTATIONS.

| Symbol | Description |
|---|---|
| $\hat{\mathbf{X}}, \hat{\mathbf{Y}}, \hat{\mathbf{Z}}$ | pre-event, post-even and regression images |
| $\mathbf{X}, \mathbf{Y}, \mathbf{Z}$ | feature matrices of $\hat{\mathbf{X}}, \hat{\mathbf{Y}}, \hat{\mathbf{Z}}$ |
| $\widetilde{\mathbf{X}}, \check{\mathbf{X}}$ | frequency representations of $\mathbf{X}$ |
| $\mathbf{X}_i$ | $i$-th row of matrix $\mathbf{X}$ |
| $x_{i,j}$ | $i$-th row and $j$-th column element of $\mathbf{X}$ |
| $G_{t1} = \{\mathcal{V}_{t1}, \mathcal{E}_{t1}, \mathbf{W}_{t1}\}$ | KNN Graph of the pre-event image |
| $G_{t2} = \{\mathcal{V}_{t2}, \mathcal{E}_{t2}, \mathbf{W}_{t2}\}$ | KNN Graph of the post-event image |
| $\mathbf{A}_{t1}, \mathbf{A}_{t2}$ | adjacent matrices |
| $\mathbf{D}_{t1}, \mathbf{D}_{t2}$ | degree matrices |
| $\mathbf{W}_{t1}, \mathbf{W}_{t2}$ | weight matrices |
| $\mathbf{P}_{t1}, \mathbf{P}_{t2}$ | random walk matrices |
| $\mathbf{L}_{t1}, \mathbf{L}_{t2}$ | Laplacian matrices |
| $\mathbf{U}, \mathbf{V}$ | orthonormal matrices of eigenvectors |
| $\mathbf{\Lambda}, \mathbf{\Gamma}$ | diagonal matrices of eigenvalues |
| $\mathbf{I}_N$ | an $N \times N$ identity matrix |
| $\mathbf{1}_N$ | an $N \times 1$ column vector of ones |

## II. PRELIMINARIES

### A. spectral filtering on graphs

Let $G = \{\mathcal{V}, \mathcal{E}, \mathbf{W}\}$ be a graph, and $\mathbf{f} = [f_1, \cdots, f_N]^T$ be the signal on the graph $G$. Different from the Fourier domain analysis for classic signal processing, the spectral representation of graph signals employs the eigenspectra (or simply "spectra" hereafter) of the graph shift operator $\mathbf{S}$ given by [10]

$$\mathbf{S} = \mathbf{U}\mathbf{\Lambda}\mathbf{U}^{-1}, \quad (1)$$

where $\mathbf{U}$ is an orthonormal matrix of the eigenvectors $\mathbf{u}_k$ in its columns and $\mathbf{\Lambda}$ is a diagonal matrix of the corresponding eigenvalues $\lambda_k$. The graph shift operator can be chosen as the adjacent matrix $\mathbf{A}$, the random walk matrix $\mathbf{P}$, or the Laplacian matrix $\mathbf{L}$, based on different applications. For the undirected graph, we have $\mathbf{S} = \mathbf{S}^T$, and $\mathbf{U}^{-1} = \mathbf{U}^T$. The graph Fourier transform (GFT) $\widetilde{\mathbf{f}}$ of a graph signal $\mathbf{f}$ is then defined as

$$\widetilde{\mathbf{f}} = \text{GFT}(\mathbf{f}) = \mathbf{U}^{-1}\mathbf{f}. \quad (2)$$

The inverse graph Fourier transform (IGFT) is defined as

$$\mathbf{f} = \text{IGFT}(\widetilde{\mathbf{f}}) = \mathbf{U}\widetilde{\mathbf{f}}. \quad (3)$$

Consider a shift-invariant system defined as the polynomials in the graph shift operator $\mathbf{S}$ of the form [10]–[12]

$$H(\mathbf{S}) = h_0\mathbf{S}^0 + h_1\mathbf{S} + \cdots + h_M\mathbf{S}^M = \sum_{m=0}^{M} h_m\mathbf{S}^m, \quad (4)$$

where $\mathbf{S}^0 = \mathbf{I}$, and $h_0, h_1, \cdots, h_M$ are system coefficients. The output $\mathbf{f}_{\text{out}}$ of the system (4) with the input signal $\mathbf{f}_{\text{in}}$ is

$$\begin{aligned}\mathbf{f}_{\text{out}} &= H(\mathbf{S})\mathbf{f}_{\text{in}} = \sum_{m=0}^{M} h_m\mathbf{S}^m\mathbf{f}_{\text{in}} = \sum_{m=0}^{M} h_m\mathbf{U}\mathbf{\Lambda}^m\mathbf{U}^{-1}\mathbf{f}_{\text{in}} \\ &= \mathbf{U}H(\mathbf{\Lambda})\mathbf{U}^{-1}\mathbf{f}_{\text{in}}.\end{aligned} \quad (5)$$

where $H(\mathbf{\Lambda}) = \sum_{m=0}^{M} h_m\mathbf{\Lambda}^m$ is the transfer function of the system. Based on (5), we have $\widetilde{\mathbf{f}}_{\text{out}} = \mathbf{U}^{-1}\mathbf{f}_{\text{out}} = H(\mathbf{\Lambda})\widetilde{\mathbf{f}}_{\text{in}}$, which is the spectral domain filtering of the graph signal.

### B. Frequency Ordering

In GSP, the frequency is defined by the eigenvalues of the graph shift $\mathbf{S}$. Specially, we define the eigenvalue decompositions as $\mathbf{L} = \mathbf{U}\mathbf{\Lambda}\mathbf{U}^{-1}$ and $\mathbf{W} = \mathbf{V}\mathbf{\Gamma}\mathbf{V}^{-1}$, with $\mathbf{\Lambda} = diag\{\lambda_1, \lambda_2, \cdots, \lambda_N\}$ and $\mathbf{\Gamma} = diag\{\gamma_1, \gamma_2, \cdots, \gamma_N\}$ representing the eigenvalues of $\mathbf{L}$ and $\mathbf{W}$, respectively.

**Definition 1.** *(Spectral ordering of the Laplacian matrix)* [13], [14]: *if we sort the spectra of Laplacian matrix $\mathbf{L}$ of the graph as $\lambda_1 \geq \lambda_2 \geq \cdots \geq \lambda_N$, then $\lambda_N$ represents the lowest frequency and $\lambda_1$ represents the highest frequency.*

**Definition 2.** *(Spectral ordering of the weighting matrix)* [12], [15]: *if we sort the spectra of weighting matrix $\mathbf{W}$ of the graph as $\gamma_1 \geq \gamma_2 \geq \cdots \geq \gamma_N$, then $\gamma_1$ represents the lowest frequency and $\gamma_N$ represents the highest frequency.*

The definitions of frequency are induced by the energy of signal change (*i.e.*, total variation), that is, we call the frequency components with smaller variations as low frequencies and call the frequency components with higher variations as high frequencies [15]. Specially, the Definition 1 is based on the 2-Dirichlet form, *i.e.*, a quadratic function as

$$\text{TV}_{\mathbf{L}}(\mathbf{f}) = \frac{1}{2}\sum_{(i,j)\in\mathcal{E}} w_{i,j}(\mathbf{f}_i - \mathbf{f}_j)^2 = \mathbf{f}^T\mathbf{L}\mathbf{f}. \quad (6)$$

Definition 2 is based on the 1-Dirichlet form total variation (TV) on a graph with signal $\mathbf{f}$ as

$$\text{TV}_{\mathbf{W}}(\mathbf{f}) = \|\mathbf{f} - \mathbf{W}_{\text{norm}}\mathbf{f}\|_1. \quad (7)$$

where the normalized weighting matrix $\mathbf{W}_{\text{norm}} = \mathbf{W}/\gamma_{\max}$ with $\gamma_{\max} = \max|\gamma_k|, k = 1, \cdots, N$.

## III. SPECTRAL PROPERTIES OF THE HETEROGENEOUS IMAGES ON DIFFERENT GRAPHS

Given two heterogeneous images of $\hat{\mathbf{X}}$ and $\hat{\mathbf{Y}}$ acquired by different sensors over the same region at different times (before and after an event), we first construct a KNN graph to represent the structure of each image as in part I [1]: $G_{t1} = \{\mathcal{V}_{t1}, \mathcal{E}_{t1}, \mathbf{W}_{t1}\}$ and $G_{t2} = \{\mathcal{V}_{t2}, \mathcal{E}_{t2}, \mathbf{W}_{t2}\}$ for pre- and post-event images, respectively. We set $\mathbf{X} = \{\mathbf{X}_1, \mathbf{X}_2, \cdots, \mathbf{X}_N\}$ and $\mathbf{Y} = \{\mathbf{Y}_1, \mathbf{Y}_2, \cdots, \mathbf{Y}_N\}$ to be the graph signals with $\mathbf{X}_i$ and $\mathbf{Y}_i$ being the feature vectors on the $i$-th vertex.

In this part, we mainly consider the case of forward regression that transforms the pre-event image to the domain of the post-event image with $\mathbf{Y} = \mathbf{Z} + \mathbf{\Delta}$, and do not consider the backward regression of transforming the post-event image to the domain of pre-event image, which is a similar process to the former. Therefore, for the sake of simplicity in defining notation, we use the eigenvalue decompositions of $\mathbf{L}_{t1}$ and $\mathbf{W}_{t1}$ in the latter part of this paper as $\mathbf{L}_{t1} = \mathbf{U}\mathbf{\Lambda}\mathbf{U}^{-1}$ and $\mathbf{W}_{t1} = \mathbf{V}\mathbf{\Gamma}\mathbf{V}^{-1}$ in descending order of eigenvalues, with $\mathbf{\Lambda} = diag\{\lambda_1, \lambda_2, \cdots, \lambda_N\}$, $\lambda_1 \geq \lambda_2 \geq \cdots \geq \lambda_N$, and $\mathbf{\Gamma} = diag\{\gamma_1, \gamma_2, \cdots, \gamma_N\}$, $\gamma_1 \geq \gamma_2 \geq \cdots \geq \gamma_N$, respectively.

### A. Low-pass property

As the frequency can be indicated by the rate of change between the vertices on the edges as illustrated by the $TV_\mathbf{L}$ and $TV_\mathbf{W}$, we can intuitively conclude that $\mathbf{X}$ is the low-pass signal on $G_{t1}$, that is, the total variations of $TV_{\mathbf{L}_{t1}}(\mathbf{X})$ and $TV_{\mathbf{W}_{t1}}(\mathbf{X})$ are very small. This can be observed in the construction process of the KNN graph, where the $i$-th vertex and $j$-th vertex are connected if and only if $i \in \mathcal{N}_j^\mathbf{x}$ (*i.e.*, $\mathbf{X}_i$ belongs to the KNN of $\mathbf{X}_j$ or $\mathbf{X}_j$ belongs to the KNN the $\mathbf{X}_i$), which means that $\mathbf{X}_i$ and $\mathbf{X}_j$ are very similar.

**Remark 1.** *The graph signal $\mathbf{X}$ is an approximate low-pass signal on the KNN graph $G_{t1}$, and the graph signal $\mathbf{Y}$ is an approximate low-pass signal on the KNN graph $G_{t2}$.*

Specifically, Remark 1 can also be demonstrated by using the Definitions 1 and 2, that is, the high frequency component of $\widetilde{\mathbf{X}}$ is almost zero. Substitute $\mathbf{L}_{t1}$ and $\mathbf{X}$ into (6), we have

$$TV_{\mathbf{L}_{t1}}(\mathbf{X}) = Tr(\mathbf{X}^T \mathbf{L}_{t1} \mathbf{X}) = Tr(\mathbf{X}^T \mathbf{U}\mathbf{\Lambda}\mathbf{U}^{-1}\mathbf{X})$$
$$= Tr(\widetilde{\mathbf{X}}^T \mathbf{\Lambda} \widetilde{\mathbf{X}}) = \sum_{k=1}^{N} \lambda_k \left\|\widetilde{\mathbf{X}}_k\right\|_2^2, \quad (8)$$

where $\widetilde{\mathbf{X}} = \mathbf{U}^{-1}\mathbf{X}$. Because $TV_{\mathbf{L}_{t1}}(\mathbf{X}) = \frac{1}{2}\sum_{(i,j)\in\mathcal{E}_{t1}} w_{i,j}^{t1} dist_{i,j}^\mathbf{x}$ is very small, then it requires that the $\left\|\widetilde{\mathbf{X}}_k\right\|_2^2$ corresponding to the large $\lambda_k$ should also be very small. According to the Definition 1 that large $\lambda_k$ represents the high frequency, we have that the high frequency component in $\widetilde{\mathbf{X}}$ is very small.

We define the normalized symmetrized weighting matrix as

$$\mathbf{W}^{sym} = \mathbf{\Phi}^{-1/2} \mathbf{W} \mathbf{\Phi}^{-1/2}, \quad (9)$$

where the diagonal matrix $\mathbf{\Phi}$ is defined such that $\mathbf{W}\mathbf{1} = \mathbf{1}$ and $\mathbf{W}^T\mathbf{1} = \mathbf{1}$. $\mathbf{W}^{sym}$ is a symmetric doubly stochastic matrix, and it can be obtained by applying the Sinkhorn-Knopp balancing algorithm [16]–[18] to iteratively normalizes the rows and columns of $\mathbf{W}$ until convergence. For the weighting matrix $\mathbf{W}^{sym}$, we have $1 = \gamma_1 \geq \gamma_2 \geq \cdots \gamma_N \geq 0$, which follows from the property of symmetric doubly stochastic matrix. We have

$$\|\mathbf{W}_{t1}^{sym}\mathbf{X}\|_F^2 = \|\mathbf{V}\mathbf{\Gamma}\mathbf{V}^{-1}\mathbf{X}\|_F^2 = Tr(\mathbf{X}^T \mathbf{V}\mathbf{\Gamma}^2\mathbf{V}^{-1}\mathbf{X})$$
$$= \sum_{k=1}^{N} \gamma_k^2 \left\|\check{\mathbf{X}}_k\right\|_2^2, \quad (10)$$

where $\check{\mathbf{X}} = \mathbf{V}^{-1}\mathbf{X}$. Due to the conservation of energy, we also have $\|\mathbf{X}\|_F^2 = \|\check{\mathbf{X}}\|_F^2 = \sum_{k=1}^N \|\check{\mathbf{X}}_k\|_2^2$. As $\mathbf{W}_{t1}^{sym}\mathbf{X} \approx \mathbf{X}$ and it requires $\sum_{k=1}^N (1 - \gamma_k^2) \|\check{\mathbf{X}}_k\|_2^2 \approx 0$, then we have $\|\check{\mathbf{X}}_k\|_2^2$

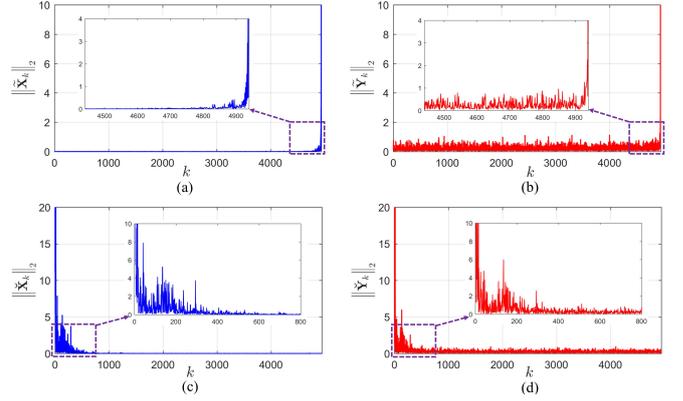

Fig. 2. Spectral properties of graph signals $\mathbf{X}$ and $\mathbf{Y}$ on the KNN graph $G_{t1}$, where $\mathbf{X}$, $\mathbf{Y}$ and $G_{t1}$ are constructed from the pre- and post-event images in Fig. 3. (a) $\left\|\widetilde{\mathbf{X}}_k\right\|_2$ with $\widetilde{\mathbf{X}} = \mathbf{U}^{-1}\mathbf{X}$; (b) $\left\|\widetilde{\mathbf{Y}}_k\right\|_2$ with $\widetilde{\mathbf{Y}} = \mathbf{U}^{-1}\mathbf{Y}$; (c) $\left\|\check{\mathbf{X}}_k\right\|_2$ with $\check{\mathbf{X}} = \mathbf{V}^{-1}\mathbf{X}$; (d) $\left\|\check{\mathbf{Y}}_k\right\|_2$ with $\check{\mathbf{Y}} = \mathbf{V}^{-1}\mathbf{Y}$.

corresponding to the small $\gamma_k$ is also very small. According to the Definitions 1 and 2 that small $\gamma_k$ represents the high frequency, we have that the high frequency component is almost zero in $\check{\mathbf{X}}$. Figure 2(a) and 2(c) show the illustration of this Remark 1.

### B. High-pass property

Next, we consider the spectral property of $\mathbf{Y}$ on the graph $G_{t1}$. As shown in Fig. 2, we find that different from the $\widetilde{\mathbf{X}}$ (or $\check{\mathbf{X}}$) on $G_{t1}$, the high frequency component of $\widetilde{\mathbf{Y}}$ (or $\check{\mathbf{Y}}$) on $G_{t1}$ is not equal to zero, which seems like the noisy component in the classic signal processing.

We decompose the $\mathbf{Y}$ into the regressed signal $\mathbf{Z}$ and changed signal $\mathbf{\Delta}$ as $\mathbf{Y} = \mathbf{Z} + \mathbf{\Delta}$, where $\mathbf{Z}$ is the assumed unchanged signal that represents the translated $\mathbf{X}$ in the domain $\mathcal{Y}$.

**Remark 2.** *The regressed signal $\mathbf{Z}$ is the low-pass signal on the KNN graph $G_{t1}$. The high frequency component of $\widetilde{\mathbf{Y}}$ (or $\check{\mathbf{Y}}$) on $G_{t1}$ is introduced by the changes caused by the event.*

First, we have that the structure of $\mathbf{Z}$ is as same as $\mathbf{X}$, that is, if $\mathbf{X}_i$ and $\mathbf{X}_j$ represent the same kind (or different kinds) of object, then $\mathbf{Z}_i$ and $\mathbf{Z}_j$ also represent the same kind (or different kinds) of object. Then, the similarity relationships of $\mathbf{Z}_i$ and $\mathbf{Z}_j$ is the same as $\mathbf{X}_i$ and $\mathbf{X}_j$. Therefore, based on the definition of frequency or by using the $TV_{\mathbf{L}_{t1}}(\mathbf{Z})$ or $TV_{\mathbf{W}_{t1}}(\mathbf{Z})$, we have that the regressed $\mathbf{Z}$ have the same spectral property as the $\mathbf{X}$ on the graph $G_{t1}$: both $\mathbf{Z}$ and $\mathbf{X}$ are the approximate low-pass signals on $G_{t1}$.

Second, Taking the GFT of the $\mathbf{Y}$ on $G_{t1}$ with Laplacian matrix $\mathbf{L}_{t1}$, we have

$$\widetilde{\mathbf{Y}} = GFT(\mathbf{Y}) = \mathbf{U}^{-1}(\mathbf{Z} + \mathbf{\Delta}) = \widetilde{\mathbf{Z}} + \widetilde{\mathbf{\Delta}}. \quad (11)$$

Because the high frequency component of $\widetilde{\mathbf{Z}}$ is approximately equal to zero, we have $\widetilde{\mathbf{Y}}_k \approx \widetilde{\mathbf{\Delta}}_k$ for high frequency $\lambda_k$, that is, the high frequency component of $\widetilde{\mathbf{Y}}$ in graph $G_{t2}$ is introduced by the changes $\mathbf{\Delta}$ caused by the event, as illustrated by the Fig. 2(b) and 2(d). Meanwhile, it should be noted that the low frequency component of $\widetilde{\boldsymbol{\Delta}}_k$ is not always equal to zero. For example, there will also be cases where adjacent vertices change at the same time, as shown in Fig. 3(c).

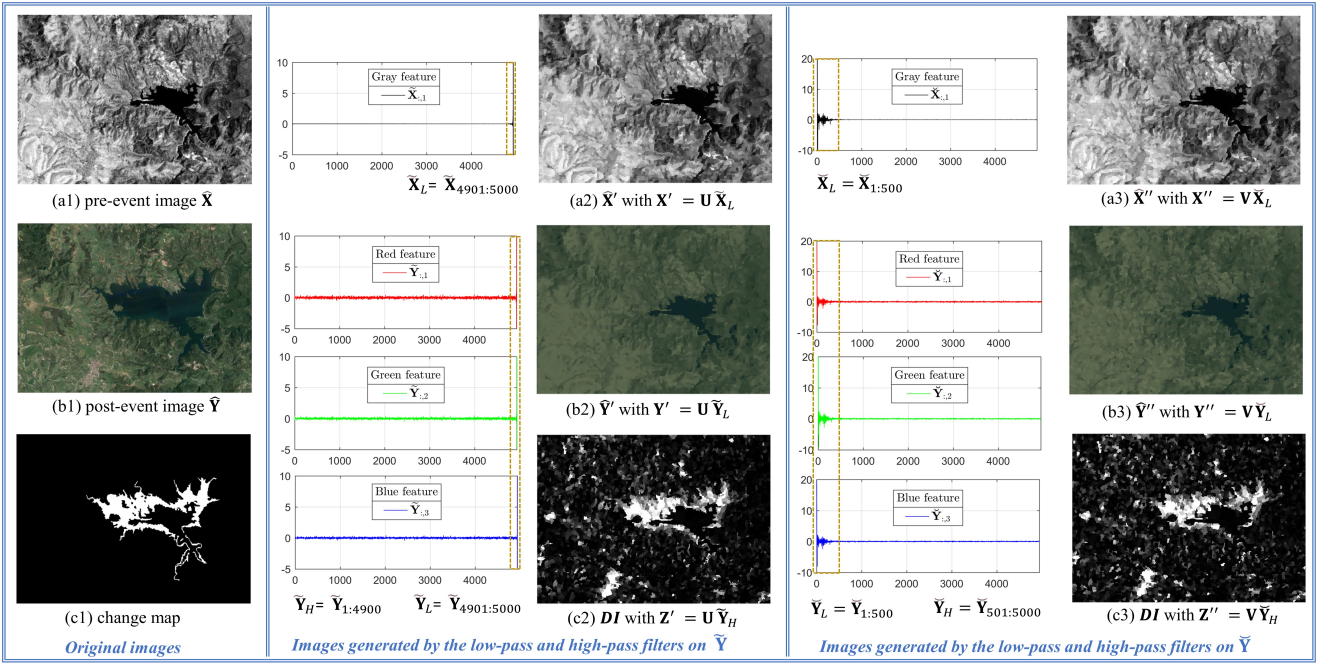

Fig. 3. Regression and changed images generated by the low-pass and high-pass filters defined on the KNN graph $G_{t1}$, respectively. By comparing (a2), (a3) with (a1), we can find that the original $\widehat{\mathbf{X}}$ is very similar to the reconstructed $\widehat{\mathbf{X}}'$ and $\widehat{\mathbf{X}}''$, with the low frequency components of $\widetilde{\mathbf{X}}_L$ and $\breve{\mathbf{X}}_L$, respectively. For the regressed $\widehat{\mathbf{Y}}'$ (b2) and $\widehat{\mathbf{Y}}''$ (b3) with the $\widetilde{\mathbf{Y}}_L$ and $\breve{\mathbf{Y}}_L$, it can be found that their structures are consistent with the pre-event image, however the appearances are similar to that of the post-event image. For the generated difference images of (c2) and (c3), we can see that the high frequency components of $\widetilde{\mathbf{Y}}_H$ and $\breve{\mathbf{Y}}_H$ can be used to detect the changes.

*C. Regression with the low frequency components*

If we divide the $\widetilde{\mathbf{Y}}$ into the low frequency and high frequency parts as $\widetilde{\mathbf{Y}} = \widetilde{\mathbf{Y}}_L + \widetilde{\mathbf{Y}}_H$ by using an ideal low-pass filter with cut-off eigenvalue $\lambda_{k_c}$, *i.e.*,

$$\left(\widetilde{\mathbf{Y}}_L\right)_k = \begin{cases} \widetilde{\mathbf{Y}}_k; & \text{if } k \geq k_c \\ 0; & \text{else} \end{cases},$$
$$\left(\widetilde{\mathbf{Y}}_H\right)_k = \begin{cases} \widetilde{\mathbf{Y}}_k; & \text{if } k < k_c \\ 0; & \text{else} \end{cases}. \quad (12)$$

With this low-pass filter, we can also divide $\widetilde{\mathbf{Z}}$ and $\widetilde{\boldsymbol{\Delta}}$ as $\widetilde{\mathbf{Z}} = \widetilde{\mathbf{Z}}_L + \widetilde{\mathbf{Z}}_H$ and $\widetilde{\boldsymbol{\Delta}} = \widetilde{\boldsymbol{\Delta}}_L + \widetilde{\boldsymbol{\Delta}}_H$, respectively. Based on the $\widetilde{\mathbf{Y}}_H \approx \widetilde{\boldsymbol{\Delta}}_H$, we have

$$\begin{aligned} \mathbf{Z} &= \text{IGFT}(\widetilde{\mathbf{Z}}) \approx \mathbf{U}\widetilde{\mathbf{Y}}_L - \mathbf{U}\widetilde{\boldsymbol{\Delta}}_L, \\ \boldsymbol{\Delta} &= \text{IGFT}(\widetilde{\boldsymbol{\Delta}}) \approx \mathbf{U}\widetilde{\mathbf{Y}}_H + \mathbf{U}\widetilde{\boldsymbol{\Delta}}_L. \end{aligned} \quad (13)$$

If we ignore the low frequency component of $\widetilde{\boldsymbol{\Delta}}_L$, we can obtain an approximate $\mathbf{Z}'$ and $\boldsymbol{\Delta}'$ as: $\mathbf{Z}' = \mathbf{U}\widetilde{\mathbf{Y}}_L$ and $\boldsymbol{\Delta}' = \mathbf{U}\widetilde{\mathbf{Y}}_H$. Figure 3 shows the approximate regression image and the changed image.

## IV. REGRESSION BASED MODEL

From the decomposition model $\mathbf{Y} = \mathbf{Z} + \boldsymbol{\Delta}$, we have that it requires the regressed $\mathbf{Z}$ have the same spectral property (*i.e.*, low-pass) as $\mathbf{X}$ in the KNN graph $G_{t1}$, which can be used as a constraint for the $\mathbf{Z}$. On the other hand, for the changed signal $\boldsymbol{\Delta}$, we have a change prior based sparse constraint for change detection problem, which is based on the fact that only a small part of the area changed and most areas remain unchanged during the event in practice. Therefore, we can obtain a decomposition based regression model for HCD problem

$$\min_{\mathbf{Z},\boldsymbol{\Delta}} g(\mathbf{Z}) + \alpha f(\boldsymbol{\Delta}) \quad s.t. \ \mathbf{Y} = \mathbf{Z} + \boldsymbol{\Delta}, \quad (14)$$

where $g(\mathbf{Z})$ represents the spectral constraint for $\mathbf{Z}$, and $f(\boldsymbol{\Delta})$ represents the prior regularization for $\boldsymbol{\Delta}$, and $\alpha > 0$ is a balancing parameter.

First, for the spectral constraint $g(\mathbf{Z})$, we have different choices. For example, we can constrain $\widetilde{\mathbf{Z}}_H \approx \mathbf{0}$ by using the penalties: 1) $g(\mathbf{Z}) = \left\|\mathbf{U}_H^{-1}\mathbf{Z}\right\|_F^2$, where $\mathbf{U}_H^{-1} = \left(\mathbf{U}^{-1}\right)_{1:k_c,:}$ represents the high-frequency transformation matrix for $\mathbf{L}_{t1}$; 2) $g(\mathbf{Z}) = \left\|\mathbf{V}_H^{-1}\mathbf{Z}\right\|_F^2$, where $\mathbf{V}_H^{-1} = \left(\mathbf{V}^{-1}\right)_{N-k_c:N,:}$ represents the high-frequency transformation matrix for $\mathbf{W}_{t1}$. Here, $\mathbf{U}_H^{-1}$ and $\mathbf{V}_H^{-1}$ are the ideal high-pass filters with the cut-off eigenvalues $\lambda_{k_c}$ and $\gamma_{k_c}$, respectively.

Although the penalty with ideal high-pass filter is intuitive, it has two drawbacks: first, it requires the eigenvalue decomposition for the graph shift operator (*e.g.*, $\mathbf{L}_{t1}$ or $\mathbf{W}_{t1}$), which requires a high computational complexity for large scale graph; second, it requires the selection of cut-off eigenvalue $\lambda_{k_c}$ or $\gamma_{k_c}$, *i.e.*, the cut-off frequency of the high-pass filter.

Based on the fact that the high frequency component of $\widetilde{\mathbf{Z}}$ almost equal to zero, we have that the value of $\sum_{k=1}^{N}\sum_{m=1}^{M} h_m \lambda_k^m \left\|\widetilde{\mathbf{Z}}_k\right\|_2^2$ is very small: first, for the small

$k$, although $\lambda_k \neq 0$, however, $\widetilde{\mathbf{Z}}_k \to \mathbf{0}$; second, for the large $k$, although $\widetilde{\mathbf{Z}}_k \neq \mathbf{0}$, however, $\lambda_k \to 0$ with $\lambda_N = 0$. Therefore, we can set $g(\mathbf{z})$ as

$$g(\mathbf{Z}) = \sum_{k=1}^{N}\sum_{m=1}^{M} h_m \lambda_k^m \left\|\widetilde{\mathbf{Z}}_k\right\|_2^2 = \sum_{m=1}^{M} h_m \text{Tr}\left(\widetilde{\mathbf{Z}}^T \mathbf{\Lambda}^m \widetilde{\mathbf{Z}}\right)$$
$$= \text{Tr}\left(\sum_{m=1}^{M} h_m \mathbf{Z}^T \mathbf{L}_{t1}^m \mathbf{Z}\right) = \text{Tr}\left(\mathbf{Z}^T H(\mathbf{L}_{t1}) \mathbf{Z}\right), \quad (15)$$

where $H(\mathbf{L}_{t1}) = \sum_{m=1}^{M} h_m \mathbf{L}_{t1}^m$. When $M = 1$, $g(\mathbf{Z})$ degenerates to the $h_1 \text{TV}_{\mathbf{L}_{t1}}(\mathbf{Z})$.

Second, for the sparsity regularization of $f(\mathbf{\Delta})$, it means that $\mathbf{\Delta}$ only exits on a small part of vertices and remains zero on other vertices. Therefore, it requires that the number of non-zero rows in $\mathbf{\Delta}$, *i.e.*, $\|\mathbf{\Delta}\|_{2,0}$, is very small. To meet this requirement, $f(\mathbf{\Delta})$ can be chosen in different forms, such as the $\ell_{2,0}$-norm [19], the $\ell_{2,1}$-norm [20], the $\ell_{2,p}$-norm with $p \in (0, 1)$ [21], and the $\ell_{\infty,1}$-norm [22], [23].

Combining the spectral constraint $g(\mathbf{Z})$, and the sparsity regularization $f(\mathbf{\Delta})$, we have the decomposition based regression model as follows

$$\min_{\mathbf{Z},\mathbf{\Delta}} \text{Tr}\left(\mathbf{Z}^T H(\mathbf{L}_{t1}) \mathbf{Z}\right) + \alpha f(\mathbf{\Delta}) \ \ s.t. \ \mathbf{Y} = \mathbf{Z} + \mathbf{\Delta}. \quad (16)$$

### A. Optimization

By using the alternating direction method of multipliers (ADMM), the augmented Lagrangian function of (16) can be written as

$$\mathcal{L}(\mathbf{Z}, \mathbf{\Delta}, \mathbf{R}) = \text{Tr}\left(\mathbf{Z}^T H(\mathbf{L}_{t1}) \mathbf{Z}\right) + \text{Tr}\left(\mathbf{R}^T (\mathbf{Y} - \mathbf{Z} - \mathbf{\Delta})\right)$$
$$+ \frac{\mu}{2} \|\mathbf{Y} - \mathbf{Z} - \mathbf{\Delta}\|_F^2 + \alpha f(\mathbf{\Delta}), \quad (17)$$

where $\mathbf{R} \in \mathbb{R}^{N \times M_y}$ is a Lagrange multiplier, and $\mu > 0$ is a penalty parameter. The minimization problem of (17) can be solved by the alternating direction method, which iteratively updates one variable at a time and fixes the others.

**Z-subproblem**. Given the current points $(\mathbf{Z}^t, \mathbf{\Delta}^t, \mathbf{R}^t)$ at the $t$-th iteration, the minimization of (17) with respect to $\mathbf{Z}$ can be formulated as

$$\mathbf{Z}^{t+1} = \arg\min_{\mathbf{Z}} \left\{ \text{Tr}\left(\mathbf{Z}^T H(\mathbf{L}_{t1}) \mathbf{Z}\right) - \text{Tr}\left((\mathbf{R}^t)^T \mathbf{Z}\right) + \frac{\mu}{2} \|\mathbf{Z} - \mathbf{Y} + \mathbf{\Delta}^t\|_F^2 \right\}. \quad (18)$$

It can be solved by taking the first-order derivative of objective function to zero, then $\mathbf{Z}$ can be updated by

$$\mathbf{Z}^{t+1} = (2H(\mathbf{L}_{t1}) + \mu \mathbf{I}_N)^{-1} \left(\mu \mathbf{Y} - \mu \mathbf{\Delta}^t + \mathbf{R}^t\right), \quad (19)$$

where $\mathbf{I}_N \in \mathbb{R}^{N \times N}$ represents an identity matrix.

**$\mathbf{\Delta}$-subproblem**. Give the fixed points $(\mathbf{Z}^{t+1}, \mathbf{\Delta}^t, \mathbf{R}^t)$, the minimization of (17) with respect to $\mathbf{\Delta}$ can be formulated as

$$\mathbf{\Delta}^{t+1} = \arg\min_{\mathbf{\Delta}} \left\{ \alpha f(\mathbf{\Delta}) - \text{Tr}\left((\mathbf{R}^t)^T \mathbf{\Delta}\right) + \frac{\mu}{2} \|\mathbf{\Delta} + \mathbf{Z}^{t+1} - \mathbf{Y}\|_F^2 \right\}, \quad (20)$$

which can be solved by the proximal operator as

$$\mathbf{\Delta}^{t+1} = prox_{\frac{\alpha}{\mu} f}\left(\mathbf{Q}^{t+1}\right), \quad (21)$$

TABLE II
IMPLEMENTATION STEPS OF ALGORITHM 1.

| Algorithm 1. Signal decomposition based regression model. |
| --- |
| **Input:** Signal $\mathbf{Y}$, graph $G_{t1}$, parameters of $\alpha, \mu, \xi^0$. |
| **Initialize:** Set $\mathbf{\Delta}^0, \mathbf{R}^0 = \mathbf{0}$, and calculate $H(\mathbf{L}_{t1})$. |
| **Repeat:** |
|   1: Update $\mathbf{Z}$ according to (19). |
|   2: Update $\mathbf{\Delta}$ according to (23), (24), or (25). |
|   3: Update $\mathbf{R}$ according to (26). |
| Until stopping criterion is met. |
| **Output:** The regressed signal $\mathbf{Z}$ and changed signal $\mathbf{\Delta}$. |

with $\mathbf{Q}^{t+1} = \mathbf{Y} - \mathbf{Z}^{t+1} + \frac{\mathbf{R}^t}{\mu}$, and the proximal operator defined as

$$prox_{\beta f}(\mathbf{b}) = \arg\min_{\mathbf{x}} f(\mathbf{x}) + \frac{1}{2\beta} \|\mathbf{x} - \mathbf{b}\|_F^2. \quad (22)$$

Depending on different regularization forms of $f(\mathbf{\Delta})$, we have different closed-form solutions for updating $\mathbf{\Delta}^{t+1}$.

If we choose $f(\mathbf{\Delta}) = \|\mathbf{\Delta}\|_{2,0}$, we have

$$\mathbf{\Delta}_i^{t+1} = \begin{cases} \mathbf{0}, & \text{if } \|\mathbf{Q}_i^{t+1}\|_2^2 \leq \frac{2\alpha}{\mu} \\ \mathbf{Q}_i^{t+1}, & \text{otherwise} \end{cases}. \quad (23)$$

If we choose the $\ell_{2,1}$-norm of $f(\mathbf{\Delta}) = \|\mathbf{\Delta}\|_{2,1}$, which is a convex relaxation of $\|\mathbf{\Delta}\|_{2,0}$, the closed-form solution of (22) can obtained by using the Lemma 3.3 of [24]

$$\mathbf{\Delta}_i^{t+1} = \max\left\{\|\mathbf{Q}_i^{t+1}\|_2 - \frac{\alpha}{\mu}\right\} \frac{\mathbf{Q}_i^{t+1}}{\|\mathbf{Q}_i^{t+1}\|_2}, \quad (24)$$

where we follow the convention $0 \cdot (0/0) = 0$.

If we have known in prior the size of changed regions, *i.e.*, the row sparsity level of $\mathbf{\Delta}$, we can construct a forced constraint as $f(\mathbf{\Delta}) = \begin{cases} 0, & \text{if } \|\mathbf{\Delta}\|_{2,0} \leq \tau \\ \infty, & \text{otherwise} \end{cases}$, and then $\mathbf{\Delta}^{t+1}$ can be updated by the hard thresholding operator as

$$\mathbf{\Delta}_i^{t+1} = \begin{cases} \mathbf{Q}_i^{t+1}, & \text{if } i \in p^\tau \\ \mathbf{0}, & \text{otherwise} \end{cases}, \quad (25)$$

where $p^\tau$ is the top $\tau$ values' indices vector of $\{\|\mathbf{Q}_i^{t+1}\|_2 | i = 1, \cdots, N_S\}$ with descending order.

**Multiplier updating**. Finally, with the fixed points $(\mathbf{Z}^{t+1}, \mathbf{\Delta}^{t+1}, \mathbf{R}^t)$, the Lagrangian multiplier can be updated as

$$\mathbf{R}^{t+1} = \mathbf{R}^t + \mu\left(\mathbf{Y} - \mathbf{Z}^{t+1} - \mathbf{\Delta}^{t+1}\right). \quad (26)$$

The procedure of solving the problem (16) is summarized in Algorithm 1 of Table II. The algorithm terminates when the maximal number of iterations is reached or the relative difference between two iteration results $\xi^{t+1} < \xi^0$, where $\xi^{t+1} = \frac{\|\mathbf{\Delta}^{t+1} - \mathbf{\Delta}^t\|_F}{\|\mathbf{\Delta}^t\|_F}$.

### B. DI and CM calculation

Once the regressed signal is computed from Algorithm 1, the regression image $\hat{\mathbf{Z}}$ can be obtained by extracting the pixel value in $\mathbf{Z}$ when $G_{t2}$ is a patch-wise graph or extracting the mean features in $\mathbf{Z}$ when $G_{t2}$ is a superpixel-wise graph. With the changed signal $\mathbf{\Delta}$ output by Algorithm 1, we can obtain the DI as $DI_{m,n} = \left(\sum_{c=1}^{C_y} \left(\hat{y}(m,n,c) - \hat{z}(m,n,c)\right)^2\right)^{1/2}$ for patch-wise graph and $DI_{m,n} = \|\boldsymbol{\Delta}_i\|_2, (m,n) \in \Omega_i$ for superpixel-wise graph.

Once the DI is obtained, the binary CM solution can be regarded as an image segmentation problem by using thresholding method or clustering method, such as the Otsu threshold method [25], K-means clustering [26], fuzzy c-means (FCM) clustering [27], or the Markov random field (MRF) based segmentation method [28].

## V. DISCUSSION

### A. Why the VDF-HCD in part I works

In the part I of this paper, we measure the change level by using the vertex domain filtering as

$$\mathbf{d^y} = H(\mathbf{S}_{t1})\mathbf{Y} - H(\mathbf{S}_{t2})\mathbf{Y}. \quad (27)$$

If we choose $H(\mathbf{S}) = \mathbf{L}$, we have $\mathbf{L}_{t1}\mathbf{Y} = \mathbf{U}\boldsymbol{\Lambda}\mathbf{U}^{-1}\mathbf{Y}$. The transfer function of $\mathbf{L}_{t1}$ is $H(\boldsymbol{\Lambda}) = \boldsymbol{\Lambda}$ with $h(\lambda_k) = \lambda_k$, which is linear amplification function as shown in the Fig. 4(c) of part I. $H(\boldsymbol{\Lambda}) = \boldsymbol{\Lambda}$ gives a large value for high frequency (*i.e.*, large eigenvalue as Definition 1), then it acts as a high-pass filter for the graph signal. Substitute the signal decomposition model of $\mathbf{Y} = \mathbf{Z} + \boldsymbol{\Delta}$, we have GFT $(\mathbf{L}_{t1}\mathbf{Y}) = \boldsymbol{\Lambda}\left(\widetilde{\mathbf{Z}} + \widetilde{\boldsymbol{\Delta}}\right)$. Then, the low-frequency components of $\widetilde{\mathbf{Y}}$ are reduced by $H(\boldsymbol{\Lambda})$, and the high-frequency components of $\widetilde{\mathbf{Y}}$ (*i.e.*, $\widetilde{\boldsymbol{\Delta}}$) are amplified by $H(\boldsymbol{\Lambda})$. As $\mathbf{Y}$ is a low-pass signal on $G_{t2}$, we have $\mathbf{L}_{t2}\mathbf{Y} \approx \mathbf{0}$. Therefore, $\mathbf{d^y} = \mathbf{L}_{t1}\mathbf{Y} - \mathbf{L}_{t2}\mathbf{Y}$ mainly contains information about $\boldsymbol{\Delta}$, which can be used to measure the changes.

Similarly, if we choose $H(\mathbf{S}) = \mathbf{W}^{\text{avg}}$, we have $H(\boldsymbol{\Gamma}) = \boldsymbol{\Gamma}$, which gives a large value for low frequency (*i.e.*, large eigenvalue as Definition 2), then it acts as a low-pass filter for the graph signal. As $\mathbf{Y}$ is the low-pass signal on $G_{t2}$, we have $\mathbf{W}_{t2}^{\text{avg}}\mathbf{Y} \approx \mathbf{Y}$. With the GFT $(\mathbf{W}_{t1}^{\text{avg}}\mathbf{Y}) = \boldsymbol{\Gamma}\left(\breve{\mathbf{Z}} + \breve{\boldsymbol{\Delta}}\right)$, the high-frequency components of $\breve{\mathbf{Y}}$ (*i.e.*, $\breve{\boldsymbol{\Delta}}$) are reduced by $H(\boldsymbol{\Lambda})$. Then $\mathbf{W}_{t1}^{\text{avg}}\mathbf{Y}$ can be regarded as an approximate $\mathbf{Z}$. Therefore, $\mathbf{d^y} = \mathbf{W}_{t1}^{\text{avg}}\mathbf{Y} - \mathbf{W}_{t2}^{\text{avg}}\mathbf{Y} \approx \mathbf{Z} - \mathbf{Y}$ can be used to measure the changes. In fact, when we choose $\mathbf{L} = \mathbf{I}_N - \mathbf{W}^{\text{avg}}$, we have $H(\boldsymbol{\Gamma}) = \mathbf{I}_N - H(\boldsymbol{\Lambda})$.

### B. How to choose the filter $H(\mathbf{S})$

For the choices of the $H(\mathbf{S})$ for vertex and spectral domain methods, we need to construct a low-pass filter for $H(\mathbf{A})$ in part I, and a high-pass filter for $H(\mathbf{L})$ in part I and part II based on the above analysis,

Suppose $\Psi(\boldsymbol{\Lambda})$ is the desired graph transfer function of a filter, we need to use the $H(\mathbf{L}) = \sum_{m=1}^{M} h_m \mathbf{L}^m$ to approximate the filter. Two methods can be used for the design of this spectral domain filter [10], [14]: first, using the least-squares approximation of $\mathbf{h} = \left(\boldsymbol{\Theta}_\lambda^T\boldsymbol{\Theta}_\lambda\right)^{-1}\boldsymbol{\Theta}_\lambda^T\psi$ with $\mathbf{h} = [h_1, \cdots, h_M]^T$ being the vector of system coefficients to be estimated, $\boldsymbol{\Theta}_\lambda$ being the Vandermonde matrix form of the eigenvalues $\lambda_k$, and $\psi = [\Psi(\lambda_1), \cdots, \Psi(\lambda_N)]^T$ being the diagonal vector of $\Psi(\boldsymbol{\Lambda})$. Second, using the polynomial approximation of $M$ degree, *e.g.*, the Chebyshev polynomial series, such as the filters used in Fig. 4(c) of part I [1].

### C. KNN graph construction

From the Fig. 3, we can find that the requirements for the KNN graph is that: first, $G_{t1}$ can represent the structure of the image; second, $\mathbf{X}$ is the low-pass signal on the graph $G_{t1}$. These correspond to the two challenges of the KNN graph: the choice of $K$ and the weighting metric.

First, a very small $K$ is not appropriate, *e.g.*, $K = 1$, each vertex is connected with is nearest neighbor. In this case, although the $\mathbf{X}$ is definitely the low-pass signal on the graph $G_{t1}$, but the $G_{t1}$ is not robust. In this case, each vertex can only get information from its nearest neighbor, which will cause many unnecessary disconnected subgraphs in $G_{t1}$. Besides, this type of graph can not adequately characterize the structure of $\hat{\mathbf{X}}$, which also means that perhaps other images that differ from $\hat{\mathbf{X}}$ will also be low-pass signal in $G_{t1}$. On the other hand, a very large $K$ is also not appropriate, *e.g.*, $K = N$, the complete graph, *i.e.*, each vertex accepts information from all other vertices whether they are similar or not. In this case, we can easily find that $\mathbf{X}$ is not a low-pass signal on the graph, or it has a large pass band at least.

Here, we recommend the adaptive probability graph with the model

$$\min_{\mathbf{W}_{t1}} \sum_{i=1}^{N}\sum_{j=1}^{N} dist_{i,j}^{\mathbf{x}} w_{i,j}^{t1} + \mathcal{R}(\mathbf{W}_{t1}, \boldsymbol{\beta}) \quad (28)$$
$$s.t.\ 0 \leq w_{i,j}^{t1} \leq 1,\ \mathbf{W}_{t1}\mathbf{1}_N = \mathbf{1}_N,$$

where $dist_{i,j}^{\mathbf{x}} = \|\mathbf{X}_i - \mathbf{X}_j\|_2^2$, and the regularization of $\mathcal{R}(\mathbf{W}_{t1}, \boldsymbol{\beta})$ aims to make the graph smooth and avoids the trivial solution. For example, if we ignore this $\mathcal{R}(\mathbf{W}_{t1}, \boldsymbol{\beta})$, then each vertex only connects with its nearest vertex with probability 1.

We present two different regularizations for $\mathbf{W}$: 1) the weighted $\ell_2$-norm as $\mathcal{R}(\mathbf{W}_{t1}, \boldsymbol{\beta}) = \sum_{i=1}^{N}\sum_{j=1}^{N}\beta_i\left(w_{i,j}^{t1}\right)^2$ used in [28], which uses the parameters $\beta_i$ to determine the number of neighbors $k_i$ of each vertex based on $k$-selection strategy. Therefore, we can find that $G_{t1}$ is a data-dependent KNN graph with adaptive neighbors selection and adaptive weight calculation. 2) The entropy regularizer of $\mathcal{R}(\mathbf{W}_{t1}, \boldsymbol{\beta}) = \sum_{i=1}^{N}\sum_{j=1}^{N}\beta_i w_{i,j}^{t1}\log w_{i,j}^{t1}$ similar as [29], which attempts to maximize the information entropy of $\mathbf{W}_{t1}$ and uses the parameters $\beta_i$ to adjust the distribution of weights. In particular, we can find that $G_{t1}$ constructed by (28) is very suitable for the proposed regression model (16): the regularization term $\sum_{i=1}^{N}\sum_{j=1}^{N} dist_{i,j}^{\mathbf{x}} w_{i,j}^{t1} = 2\text{Tr}\left(\mathbf{X}^T \mathbf{L}_{t1}\mathbf{X}\right)$ in the objective function of graph construction model (28) is consistent with the penalty term $\text{Tr}\left(\mathbf{Z}^T H(\mathbf{L}_{t1})\mathbf{Z}\right)$ in the regression model.

### D. Extended graphs

In this paper, we construct the KNN graph for each image to capture the structure information. An important reason why the KNN graph was chosen is that it can distinguish between the changed and unchanged signals in the spectral domain: the original $\mathbf{X}$ and regressed $\mathbf{Z}$ are low-pass signals in the KNN graph $G_{t1}$, while the high-frequency components of $\widetilde{\mathbf{Y}}$ are introduced by the changed signal $\boldsymbol{\Delta}$. Similarly, if we can find such a graph $G'$ that the changed and unchanged signals can be discriminated in the spectral domain with $G'$, then more regularization terms $g(\mathbf{Z})$ can be added to the regression model (14).

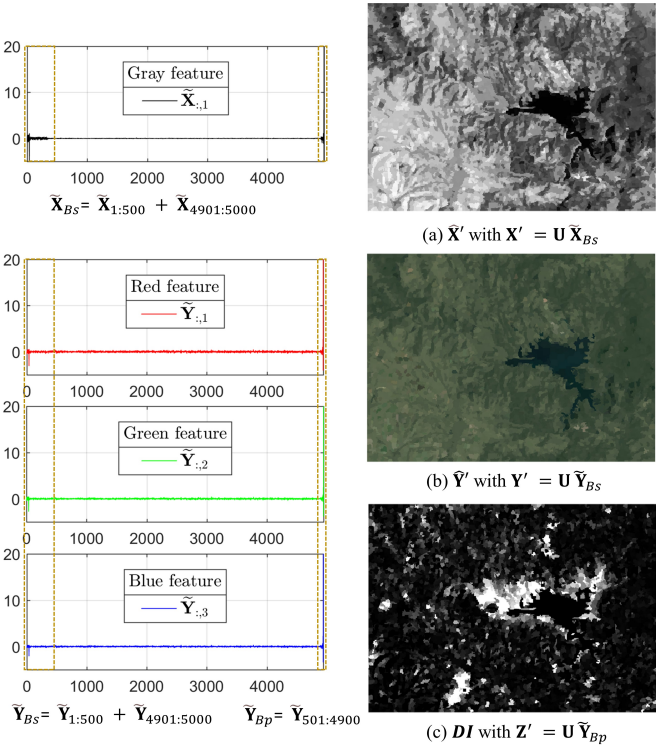

Fig. 4. Regression image (b) and difference image (c) generated by the band-stop and band-pass filters defined on the KFN graph $G'_{t1}$ respectively, where $\mathbf{X}$, $\mathbf{Y}$ and $G'_{t1}$ are constructed from the pre- and post-event images in Fig. 3. The high-frequency component of $\widetilde{\mathbf{X}}$ in the graph $G'_{t1}$, i.e., $\widetilde{\mathbf{X}}_{4901:5000}$ is large. The approximate regressed image can be obtained with partial low and high frequency components, i.e., $\widetilde{\mathbf{Y}}_{Bs}$, and the difference image can be calculated with partial mid-frequency components, i.e., $\widetilde{\mathbf{Y}}_{Bp}$.

If we construct a K-farthest neighbors (KFN) graph $G'_{t1} = \{\mathcal{V}'_{t1}, \mathcal{E}'_{t1}, \mathbf{W}'_{t1}\}$ for the pre-event image, where each vertex is connected with its K-farthest neighbors, i.e., $(i,j) \in \mathcal{E}'_{t1}$ if and only if $dist^{\mathbf{x}}_{i,j}$ is among the $K$-largest elements of the vector $dist^{\mathbf{x}}_i$ or the vector $dist^{\mathbf{x}}_j$. In this way, the change rate of signal $\mathbf{X}$ between the vertices on the edges of $G'_{t1}$ is very rapid, that is, the total variation of $\mathrm{TV}_{\mathbf{L}'_{t1}}(\mathbf{X})$ is very large. Therefore, the high-frequency component of $\widetilde{\mathbf{X}}$ in the graph $G'_{t1}$ is also large, so as to the regression signal $\mathbf{Z}$, as shown in Fig. 4. To exploit this property, we can add a KFN graph induced repulsive regularization term $g(\mathbf{Z}) = -\mathrm{TV}_{\mathbf{L}'_{t1}}(\mathbf{Z})$ or $g(\mathbf{Z}) = \sum_{(i,j) \in \mathcal{E}'_{t1}} w^{t1'}_{i,j} \exp\left(-dist^{\mathbf{z}}_{i,j}\right)$ in the regression model (14), which requires neighboring nodes of $\mathbf{Z}$ connected by the KFN graph $G'_{t1}$ to be further apart (because they do not represent the same type of object). Figure 4 shows an example of the usage of KFN in HCD. Similarly, other types of graph properties can be extended in the proposed GSP based HCD framework.

### E. Framework of GSP-SDA based CD

Here, we give a general framework for CD problem (with both homogeneous and heterogeneous multi-temporal remote sensing images) based on the GSP with spectral domain analysis (SDA).

Step 1. Construct the graph and graph signals.

Step 2. Chose the penalties of $g(\mathbf{Z})$ and $f(\mathbf{\Delta})$ to obtain the regression model of (14).

Step 3. Solve the minimization problem to obtain the changed signal.

Step 4. Segment the DI to obtain the final CM.

In the step 1, it requires that the graph signals are distinguishable in the spectral domain of the graph. In the step 2, the $g(\mathbf{Z})$ is corresponding to the spectral property of the signal, such as the low-pass property of the KNN graph; the $f(\mathbf{\Delta})$ can also be other prior knowledge based penalties in addition to the sparsity penalty, such as the low-rank, smoothness, and some known labels of the semi-supervised (or supervised) CD problem. In the step 4, there are many existing methods that can be used for this binary classification problem.

## VI. EXPERIMENTS

In this section, we analyze the performance of the proposed SDA-HCD. We first provide a a brief description of the experimental setting, then present the experimental results on different HCD data sets, and finally some detailed discussions are made.

### A. Experimental setting

We use the same heterogeneous data sets (Datasets #1-#7 as listed in Table III of part I) and evaluation metrics as in part I [1], e.g., the receiver operating characteristic (ROC) and precision-recall (PR) curves, the areas under the ROC curve (AUR) and PR curve (AUP), overall accuracy (OA), Kappa coefficient (Kc), and F1-measure (Fm). For the SDA-HCD, we construct the superpixel-wise graph with $N = 10^4$, use the mean, median, and variance values of superpixel to construct the graph signals of $\mathbf{X}$ and $\mathbf{Y}$, and choose the adaptive model (28) with weighted $\ell_2$-norm to construct the graph $G_{t1}$. We choose $\ell_{2,1}$-norm of $f(\mathbf{\Delta}) = \|\mathbf{\Delta}\|_{2,1}$ and set the balancing parameter $\alpha = 0.05$ for the regression model (14). We choose the $H(\mathbf{L}_{t1}) = \mathbf{L}_{t1} + h_2\mathbf{L}^2_{t1} + h_3\mathbf{L}^3_{t1}$ with $h_2 = h_3 = 1$ for the results in section VI-B, and leave the discussion of $H(\mathbf{L}_{t1})$ in section VI-C.

### B. Experimental results

We compare the proposed method with following regression based HCD methods on all the heterogeneous data sets.

HPT [2]. A supervised kernel regression method that estimates the mapping pixels using known unchanged pixels estimated mapping pixels using known unchanged pixels. In the experiments, we use 40% of the unchanged pixels as training samples.

AMD-IR [3]. An unsupervised regression method uses the AMD to pick probably unchanged pixels as the pseudo-training set.

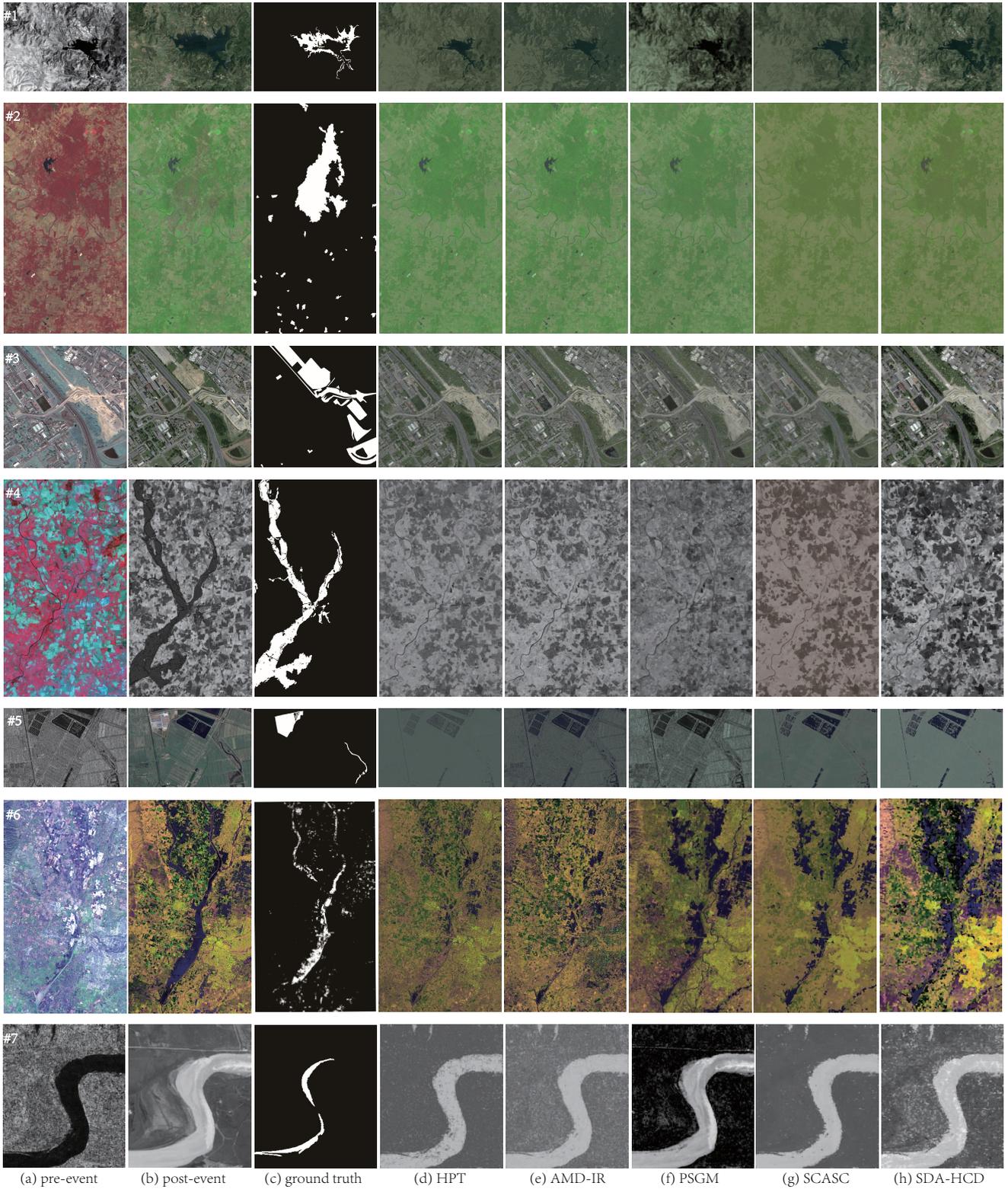

Fig. 5. Regression images generated by different methods on all the heterogeneous data sets. From top to bottom, they correspond to Datasets #1 to #7, respectively. From left to right are: (a) pre-event image; (b) post-event image; (c) ground truth; (d) regression image of HPT; (e) regression image of AMD-IR; (f) regression image of PSGM; (g) regression image of SCASC; (h) regression image of SDA-HCD.

TABLE III
AUR AND AUP OF DIS GENERATED BY DIFFERENT METHODS ON THE HETEROGENEOUS DATA SETS. THE HIGHEST SCORES ARE HIGHLIGHTED IN BOLD.

| Measures | Methods | #1 | #2 | #3 | #4 | #5 | #6 | #7 | Average |
|---|---|---|---|---|---|---|---|---|---|
| AUR | HPT [2] | **0.889** | 0.920 | 0.770 | **0.947** | 0.911 | **0.928** | 0.964 | 0.904 |
|  | AMD-IR [3] | 0.795 | 0.918 | 0.708 | 0.927 | 0.922 | 0.867 | 0.859 | 0.857 |
|  | PSGM [30] | 0.837 | 0.898 | 0.777 | 0.933 | **0.975** | 0.911 | 0.728 | 0.866 |
|  | SCASC [28] | 0.885 | **0.932** | 0.793 | 0.936 | 0.968 | 0.887 | **0.969** | **0.910** |
|  | SDA-HCD (proposed) | **0.889** | 0.854 | **0.824** | 0.928 | 0.958 | 0.871 | 0.956 | 0.897 |
| AUP | HPT [2] | 0.373 | **0.575** | 0.355 | **0.806** | 0.470 | **0.506** | 0.602 | 0.527 |
|  | AMD-IR [3] | 0.155 | 0.536 | 0.264 | 0.741 | 0.564 | 0.289 | 0.216 | 0.395 |
|  | PSGM [30] | **0.593** | 0.526 | 0.405 | 0.745 | **0.794** | 0.484 | 0.152 | 0.528 |
|  | SCASC [28] | 0.383 | 0.566 | 0.458 | 0.636 | 0.695 | 0.447 | 0.597 | 0.540 |
|  | SDA-HCD (proposed) | 0.457 | 0.565 | **0.540** | 0.646 | **0.794** | 0.434 | **0.680** | **0.588** |

PSGM [30]. An unsupervised method that learns a self-expression based patch similarity graph matrix (PSGM) for one image, then computes the regression image by multiplying the other image with this learned matrix.

SCASC [28]. An unsupervised method that uses the regression model with sparse constrained adaptive structure consistency (SCASC), which is similar to the proposed SDA-HCD but without using the high-order information and spectral analysis.

*1) Image regression:* In the first experiment, we verify the effectiveness of the SDA based image regression model. Figure 5 shows the regression images generated by different methods on all the heterogeneous data sets, which transforms the pre-event image to the domain of post-event image. Intuitively, these methods basically complete the image translation. That is, in the unchanged area, the regression image and the post-event image are similar, while they appear very different in the changed area, which can be used to detect the changes. Visually, we can see that SDA-HCD performs better on Datasets #1, #3 and #7; PSGM performs better on Dataset #5 and worse on Dataset #7; HPT performs better on Datasets #4 and #6. Also, we can find that the superpixel-based regression method have a block smoothing effect, such as the regression images of SCASC and SDA-HCD on Datasets #4 and #6.

*2) Difference image:* In order to evaluate the ability of the algorithm to measure change, we show the DIs generated by different methods in Fig. 6, and plot the corresponding ROC and PR curves in Fig. 7. The AUR of ROC curves and the AUP of PR curves are listed in Table III.

In the DIs of Datasets #1 and #5, PSGM performs best, SDA-HCD comes second, as shown in Fig. 6(b)-(f) and Fig. 7. Meanwhile, we can also find that the DIs generated by PSGM, SCASC and SDA-HCD in Fig. 6 are sparse (*e.g.*, especially in the DIs of Datasets #1, #2 and #7), which is due to the prior sparsity based regularization used in these methods. In the DIs of Datasets #4 and #6, HPT performs better than other methods. However, these results are attributed to the fact that HPT employs a large amount of labeled unchanged samples (40%). SDA-HCD can obtain better DIs on the Datasets #3 and #7 than other comparison methods, as demonstrated by the ROC and PR curves in Fig. 6. Comparing SCASC and SDA-HCD, we can find that they produce similar DIs, because they both belong to the structured graph based regression methods; however, there are differences in the performance of their DIs in Fig. 7 and Table III, *i.e.*, SDA-HCD outperforms SCASC, which is due to the use of high-order information and spectral analysis in SDA-HCD. In general, as can be seen from the results of Figs. 6 and 7 and Table III, the DIs of SDA-HCD can well distinguish between changed and unchanged areas, and SDA-HCD gains the highest average AUP of 0.588, which is 4.4% higher than the second ranked SCASC.

*3) Change maps:* Fig. 6(h)-(k) show the CMs of different methods on all the evaluated data sets. Intuitively, the results generated by these comparison methods can generally reflect main information of changes. To be specific, the CMs provided by HPT, AMD-IR are affected by the salt-and-pepper noise as show in Fig. 6(h) and 6(i), especially on the Dataset #3, which is because their pixel-to-pixel regression process is not robust enough to cope with complex and variable HCD conditions. As a result, a large number of unchanged pixels are misclassified into changed ones, *e.g.*, the CMs of HPT on Datasets #3 and #5, the CMs of AMD-IR on Datasets #1, #3 and #7. The PSGM achieves relatively good performances on some data sets, such as Datasets #1 and #5, but also has higher FP on some data sets, such as Dataset #7. SCASC can achieve satisfactory CMs with less FP on Datasets #2, #4 and #7, thanks to the exploitation of graph based robust structure consistency between heterogeneous images. Nevertheless, due to the fact that SCASC ignores the higher-order neighborhood information hidden in the graphs, making its performance not as good as SDA-HCD. For example, interpreted in detail, the results of SDA-HCD are more consistent with the ground truth, especially in some regions marked by the yellow boxes on the Datasets #1 and #7 in Fig. 6(j) and 6(k).

Table IV reports the quantitative evaluation results of comparison methods on the data sets. Clearly, the quantitative results are consistent with the visual analysis of Fig. 6. As can be seen from Table IV, the SDA-HCD outperforms others on Datasets #4, #5 and #7, and obtain very competitive performance on other data sets. The average OA, Kc and Fm obtained by SDA-HCD on all the evaluated data sets are about 0.955, 0.660, and 0.683, respectively. These scores are higher than other methods, *e.g.*, the average Kc and Fm are improved by 1.7% and 1.6% respectively compared to the second ranked SCASC. This demonstrates SDA based method can efficiently

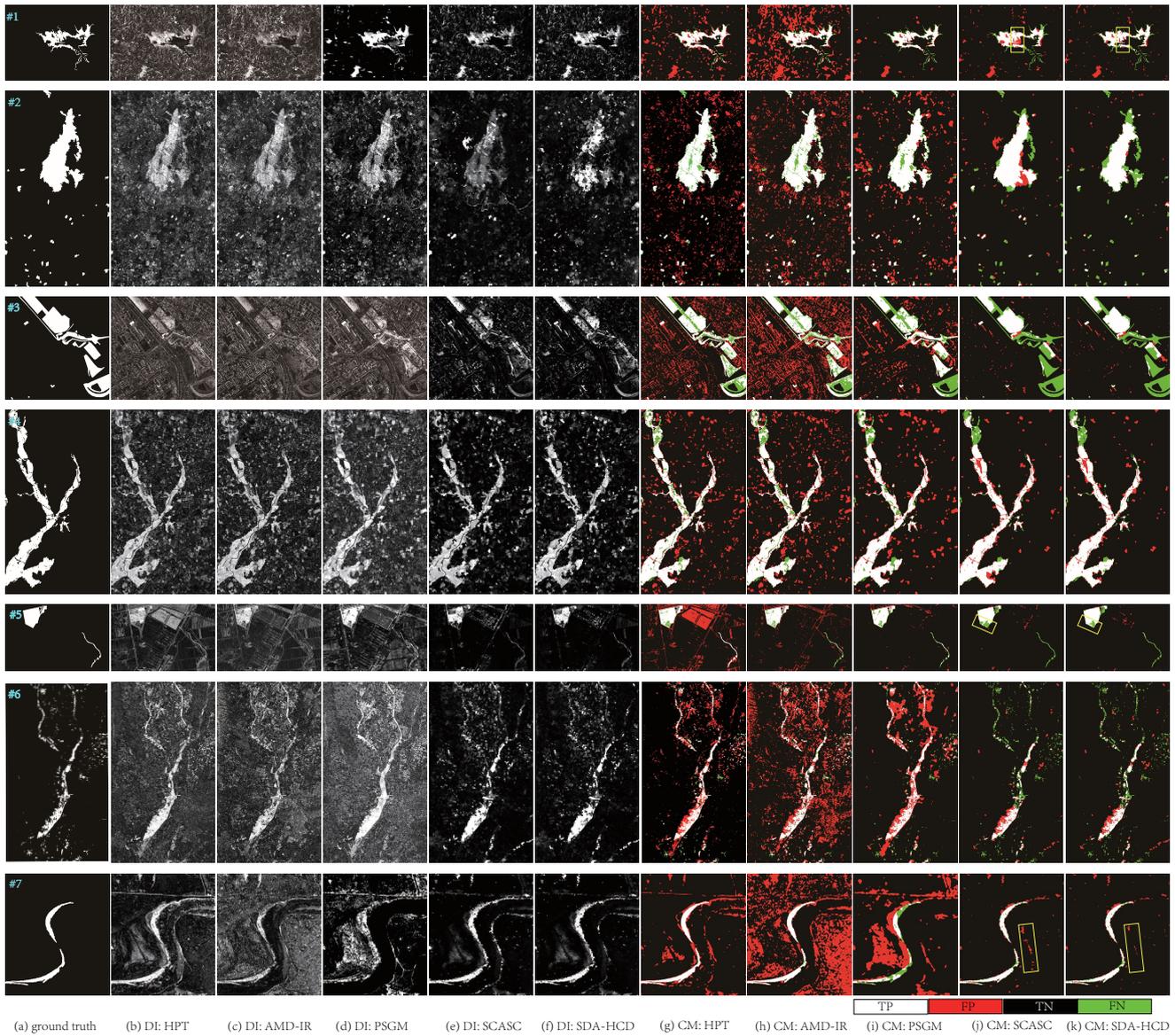

Fig. 6. DIs and CMs generated by different methods on all the heterogeneous data sets. From top to bottom, they correspond to Datasets #1 to #7, respectively. From left to right are: (a) ground truth; (b) DI of HPT; (c) DI of AMD-IR; (d) DI of PSGM; (e) DI of SCASC; (f) DI of SDA-HCD; (g) CM of HPT; (h) CM of AMD-IR; (i) CM of PSGM; (j) CM of SCASC; (k) CM of SDA-HCD. In the binary CM, White: true positives (TP); Red: false positives (FP); Black: true negatives (TN); Green: false negatives (FN).

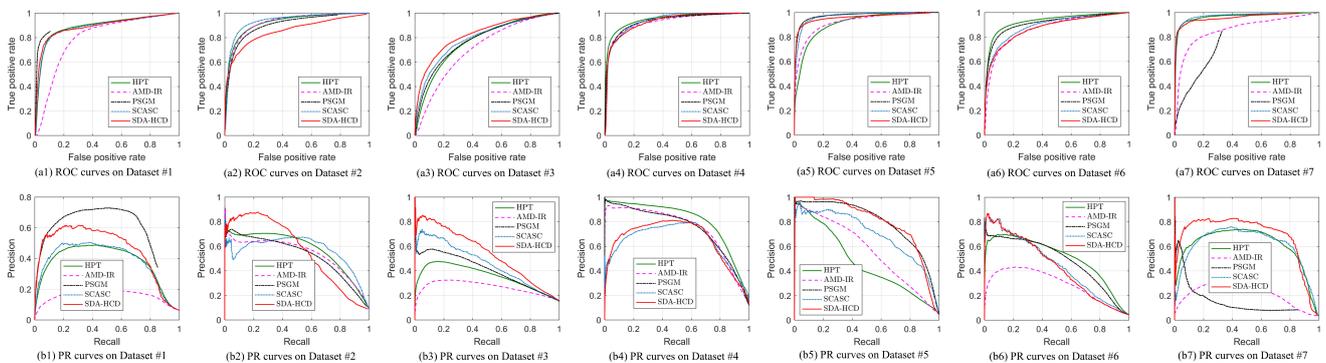

Fig. 7. ROC and PR curves of DIs generated by different methods on all the heterogeneous data sets. The top row are the ROC curves, the bottom row are the PR curves. From left to right are the results on Datasets #1 to #7, respectively.

TABLE IV
QUANTITATIVE MEASURES OF BINARY CMS ON THE HETEROGENEOUS DATA SETS. THE HIGHEST SCORES ARE HIGHLIGHTED IN BOLD.

| Datasets | HPT [2] | | | AMD-IR [3] | | | PSGM [30] | | | SCASC [28] | | | SDA-HCD (proposed) | | |
|---|---|---|---|---|---|---|---|---|---|---|---|---|---|---|---|
| | OA | Kc | Fm | OA | Kc | Fm | OA | Kc | Fm | OA | Kc | Fm | OA | Kc | Fm |
| Dataset #1 | 0.912 | 0.504 | 0.546 | 0.799 | 0.255 | 0.328 | **0.961** | **0.682** | **0.703** | 0.947 | 0.593 | 0.621 | 0.956 | 0.653 | 0.677 |
| Dataset #2 | 0.916 | 0.589 | 0.633 | 0.901 | 0.547 | 0.598 | 0.928 | 0.631 | 0.670 | 0.961 | **0.751** | **0.772** | **0.963** | 0.732 | 0.751 |
| Dataset #3 | 0.815 | 0.415 | 0.523 | 0.724 | 0.259 | 0.411 | 0.857 | 0.473 | **0.558** | 0.892 | 0.464 | 0.516 | **0.896** | **0.488** | 0.539 |
| Dataset #4 | 0.918 | 0.671 | 0.717 | 0.898 | 0.607 | 0.664 | 0.922 | 0.675 | 0.719 | 0.949 | 0.771 | 0.800 | **0.950** | **0.773** | **0.803** |
| Dataset #5 | 0.859 | 0.318 | 0.368 | 0.950 | 0.572 | 0.597 | 0.977 | 0.744 | 0.756 | 0.979 | 0.741 | 0.751 | **0.982** | **0.778** | **0.787** |
| Dataset #6 | 0.932 | **0.488** | **0.518** | 0.822 | 0.236 | 0.291 | 0.908 | 0.383 | 0.422 | **0.961** | 0.479 | 0.500 | 0.959 | 0.468 | 0.489 |
| Dataset #7 | 0.927 | 0.447 | 0.476 | 0.686 | 0.116 | 0.170 | 0.832 | 0.124 | 0.173 | 0.977 | 0.700 | 0.711 | **0.980** | **0.731** | **0.741** |
| Average | 0.897 | 0.490 | 0.540 | 0.826 | 0.370 | 0.437 | 0.912 | 0.530 | 0.572 | 0.952 | 0.643 | 0.667 | **0.955** | **0.660** | **0.683** |

TABLE V
ACCURACY RATE OF CMS GENERATED BY DIFFERENT METHODS ON DIFFERENT DATA SETS. THE RESULTS OF THESE COMPARISON METHODS ARE REPORTED BY THEIR ORIGINAL PUBLISHED PAPERS, EXCEPT THE RESULTS INDICATED WITH † ARE REPORTED BY [38] (THEY ARE CONSISTENT WITH THEIR OPEN SOURCE CODES IN [39]). ITALICIZED AND UNDERLINED MARKS ARE USED FOR DEEP LEARNING BASED METHODS.

| Dataset #1 | OA |
|---|---|
| *DFR-MT* [31] | 0.975 |
| *CACFL* [32] | 0.975 |
| VDF-HCD | **0.971** |
| ALSC [33] | 0.965 |
| SDA-HCD | **0.956** |
| MDS [34] | 0.942 |
| *AFL-DSR* [35] | 0.929 |
| RMN [36] | 0.847 |

| Dataset #2 | OA |
|---|---|
| VDF-HCD | **0.973** |
| SDA-HCD | **0.963** |
| *DCCAE* [37] | 0.957 |
| *DCCA* [38] | 0.947 |
| †KCCA [39] | 0.915 |
| †CCA [39] | 0.812 |

| Dataset #3 | OA |
|---|---|
| GIR-MRF [40] | 0.901 |
| SDA-HCD | **0.896** |
| VDF-HCD | **0.886** |
| *AFL-DSR* [35] | 0.880 |
| RMN [36] | 0.877 |
| NLPEM [41] | 0.853 |

| Dataset #4 | OA |
|---|---|
| SDA-HCD | **0.950** |
| VDF-HCD | **0.944** |
| GIR-MRF [40] | 0.936 |
| *AFL-DSR* [35] | 0.836 |
| MDER [42] | 0.818 |

| Dataset #5 | OA |
|---|---|
| *DPFL* [43] | 0.987 |
| VDF-HCD | **0.985** |
| SDA-HCD | **0.982** |
| GIR-MRF [40] | 0.982 |
| *AFL-DSR* [35] | 0.980 |
| *CACFL* [32] | 0.979 |
| MDS [34] | 0.967 |
| *LT-FL* [44] | 0.964 |
| ALSC [33] | 0.963 |
| RMN [36] | 0.884 |

| Dataset #6 | OA |
|---|---|
| SDA-HCD | **0.959** |
| GIR-MRF [40] | 0.959 |
| VDF-HCD | **0.952** |
| *DPFL* [43] | 0.945 |
| ALSC [33] | 0.944 |
| *ACE-Net* [7] | 0.915 |
| *X-Net* [7] | 0.911 |
| *SSL* [45] | 0.902 |

| Dataset #7 | OA |
|---|---|
| VDF-HCD | **0.982** |
| *LT-FT* [44] | 0.981 |
| SDA-HCD | **0.980** |
| *DPFL* [43] | 0.978 |
| *CACFL* [32] | 0.977 |
| *SCCN* [46] | 0.977 |
| *ASDNN* [47] | 0.975 |

improve the HCD performance.

Finally, in order to further compare the performance of the proposed graph signal processing based HCD methods, *i.e.*, the VDF-HCD in part I [1] and SDA-HCD in part II, the results obtained by some representative and SOTA methods [31]–[47] are summarized in Table V, except for M3CD [48], FPMS [49], CICM [50], NPSG [51] and IRG-McS [52], HPT [2], AMD-IR [3], PSGM [30], and SCASC [28], which have been compared in detail in this paper. Among these comparison approaches, *DFR-MT* [31], *CACFL* [32], *AFL-DSR*, *DCCAE* [37], *DCCA* [38], *DPFL* [43], *LT-FT* [44], *SSL* [45], *ACE-Net* [7], *X-Net* [7], *SCCN* [46], and *ASDNN* [47] are deep learning based methods. For the sake of fairness, we directly quote the results of the corresponding data sets in their original published papers in Table V. Because the datasets used in each paper are not identical, Table V is not aligned. As can be seen in Table V, the proposed graph signal processing based methods (VDF-HCD and SDA-HCD) consistently yields better or very competitive accuracy across different data sets by comparing with these SOTA approaches, which again demonstrates the effectiveness of the proposed GSP perspective for HCD problem.

### C. Discussions

*1) The choices of $H(\mathbf{L}_{t1})$:* As analyzed in Section IV, the proposed decomposition based regression model (14) requires the spectral constraint $g(\mathbf{Z})$ to penalize high frequency components of $\widetilde{\mathbf{Z}}_H$, such that the regressed $\mathbf{Z}$ have the same spectral property (low-pass) as $\mathbf{X}$ in the KNN graph $G_{t1}$, *i.e.*, $\widetilde{\mathbf{Z}}_H \approx \mathbf{0}$.

If we decompose $H(\mathbf{L}_{t1}) = H^{\frac{1}{2}}(\mathbf{L}_{t1}) H^{\frac{1}{2}}(\mathbf{L}_{t1})$, then we have $g(\mathbf{Z}) = \left\| H^{\frac{1}{2}}(\mathbf{L}_{t1}) \mathbf{Z} \right\|_F^2$, where $H^{\frac{1}{2}}(\mathbf{L}_{t1})$ represents a graph filter with the transfer function: $H^{\frac{1}{2}}(\mathbf{\Lambda}) = \left( \sum_{m=1}^M h_m \mathbf{\Lambda}^m \right)^{\frac{1}{2}}$. Therefore, it requires that $H^{\frac{1}{2}}(\mathbf{L}_{t1})$ is an approximate high-pass filter.

In Fig. 8, we plot different functions of $H^{\frac{1}{2}}(\lambda) = \left( \sum_{m=1}^M h_m \lambda^m \right)^{\frac{1}{2}}$. In these functions, we can find that the higher-order polynomial transfer function $H^{\frac{1}{2}}(\lambda) = \sqrt{\lambda + \lambda^2 + \lambda^3}$ is closer to a high-pass function than $H^{\frac{1}{2}}(\lambda) = \sqrt{\lambda}$. This also intuitively explains why SDA-HCD, which uses higher-order information and spectral analysis, performs better than the SCASC, as illustrated by the results of Figs. 5 to 7 and Tables III and IV.

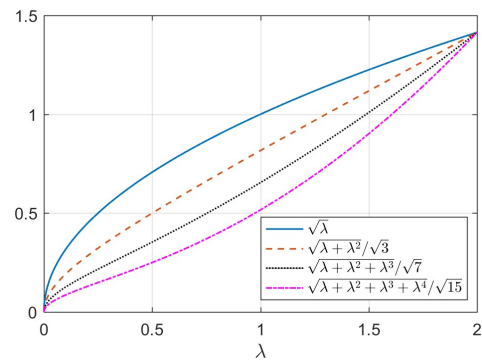

Fig. 8. Different functions of $H^{\frac{1}{2}}(\lambda) = \left( \sum_{m=1}^M h_m \lambda^m \right)^{\frac{1}{2}}$.

TABLE VI
AUR AND AUP OF DIS GENERATED BY DIFFERENT METHODS ON THE HETEROGENEOUS DATA SETS.

| SDA-HCD with $H(\mathbf{L}_{t1})$ | Average scores | | | | |
|---|---|---|---|---|---|
| | AUR | AUP | OA | Kc | Fm |
| $\mathbf{L}_{t1}$ | 0.896 | 0.533 | 0.945 | 0.633 | 0.654 |
| $\mathbf{L}_{t1} + \mathbf{L}_{t1}^2$ | 0.897 | 0.578 | 0.953 | 0.655 | 0.677 |
| $\mathbf{L}_{t1} + \mathbf{L}_{t1}^2 + \mathbf{L}_{t1}^3$ | 0.897 | 0.588 | 0.955 | 0.660 | 0.683 |
| $\mathbf{L}_{t1} + \mathbf{L}_{t1}^2 + \mathbf{L}_{t1}^3 + \mathbf{L}_{t1}^4$ | 0.896 | 0.581 | 0.955 | 0.654 | 0.679 |

TABLE VII
COMPUTATIONAL TIME (SECONDS) OF EACH PROCESS OF SDA-HCD.

| Data sets | $N$ | $t_{pre}$ | $t_{gc}$ | $t_{ir}$ | $t_{total}$ |
|---|---|---|---|---|---|
| Dataset #7 | 5000 | 0.29 | 0.12 | 5.28 | 6.84 |
| $343 \times 291 \times 1(3)$ | 10000 | 0.63 | 0.36 | 21.75 | 27.20 |
| Dataset #2 | 5000 | 2.85 | 0.13 | 5.60 | 10.22 |
| $2000 \times 2000 \times 3(3)$ | 10000 | 3.23 | 0.35 | 21.91 | 29.77 |

In Table VI, we list the quantitative results (average scores on all the evaluated data sets) of SDA-HCD with different $H(\mathbf{L}_{t1})$, including $\mathbf{L}_{t1}$, $\mathbf{L}_{t1}+\mathbf{L}_{t1}^2$, $\mathbf{L}_{t1}+\mathbf{L}_{t1}^2+\mathbf{L}_{t1}^3$, and $\mathbf{L}_{t1}+\mathbf{L}_{t1}^2+\mathbf{L}_{t1}^3+\mathbf{L}_{t1}^4$. With respect to the first-order $H(\mathbf{L}_{t1}) = \mathbf{L}_{t1}$, high-order $H(\mathbf{L}_{t1})$ improves by 4.9% and 2.6% on average AUP and Fm in Table VI, respectively. By contrast, the fourth-order $H(\mathbf{L}_{t1})$ deteriorates the performance a little bit on the average scores, which is perhaps caused by the excessively narrow bandwidth of the transfer function.

*2) Computational analysis:* The main computational complexity of the proposed method is concentrating on the graph construction and image regression. For the former, calculating the distance matrix between all the patches or superpixels requires $\mathcal{O}(M_x N^2/2)$, and sorting the distance matrix by column to construct the adaptive graph (28) with weighted $\ell_2$-norm $\mathcal{R}(\mathbf{W}_{t1}, \boldsymbol{\beta})$ requires $\mathcal{O}(N^2 \log N)$ [28]. For the image regression model of Algorithm 1, updating $\mathbf{Z}$ with (19) requires $\mathcal{O}(N^3)$, updating $\boldsymbol{\Delta}$ requires $\mathcal{O}(M_y N)$ by using the closed-form proximal operator (21), and updating $\mathbf{R}$ with (26) requires $\mathcal{O}(M_y N)$. Although the complexity of Algorithm 1 is very high, which requires $\mathcal{O}(N^3)$ for each iteration, two acceleration strategies are available to improve the efficiency of the Algorithm 1 as introduced in [28]: computing the matrix inversion of $(2H(\mathbf{L}_{t1}) + \mu \mathbf{I}_N)^{-1}$ off-line in advance or solving the $\mathbf{Z}$-subproblem (18) with preconditioned conjugate gradient (PCG) method.

Table VII reports the computational time of each process of SDA-HCD with different $N$ on Datasets #1 and #3, which is performed in MATLAB 2016a running on a Windows desktop with Intel Core i7-8700K CPU. In Table VII, $t_{pre}$, $t_{gc}$, and $t_{ir}$ represent the computational times spent in the pre-processing, graph construction, and image regression respectively, $t_{total}$ represents the total running time. As can be seen in Table VII, the running time of SDA-HCD is mainly determined by the graph scale $N$, and image regression is the most time-consuming process in SDA-HCD.

## VII. CONCLUSION

In this part, we analyze the GSP for HCD from the spectral domain. We first show the spectral properties of the heterogeneous images on the same graph, and illustrate that it is the changes between images that cause the differences in their spectral properties. Based on our finds, we propose a signal decomposition based regression model for HCD, which decomposes the source signal into the regressed signal and changed signal, and constrains the spectral properties of the regressed signal and target signal to be consistent. Experimental results on seven heterogeneous data sets demonstrate the effectiveness of the proposed SDA-HCD.

In this paper, we provide a new perspective for HCD problem and propose two methods with GSP, *i.e.*, VDF-HCD from the vertex domain and SDA-HCD from the spectral domain. However, we only consider the simple graph, linear filtering and graph Fourier transform in this paper. Our future work includes studying hypergraph, graph wavelet analysis and graph neural networks on HCD problem.

ACKNOWLEDGMENT

The authors would like to thank the researchers for their friendly sharing of heterogeneous change detection codes and data sets, which provide a wealth of resources for this study.

REFERENCES

[1] Y. Sun, L. Lei, D. Guan, G. Kuang, and L. Liu, "Graph signal processing for heterogeneous change detection—part I: Vertex domain filtering," *submitted to IEEE Transactions on Geoscience and Remote Sensing*, pp. 1–15, 2022. 1, 3, 7, 8, 12
[2] Z. Liu, G. Li, G. Mercier, Y. He, and Q. Pan, "Change detection in heterogenous remote sensing images via homogeneous pixel transformation," *IEEE Transactions on Image Processing*, vol. 27, no. 4, pp. 1822–1834, 2018. 2, 8, 10, 12
[3] L. T. Luppino, F. M. Bianchi, G. Moser, and S. N. Anfinsen, "Unsupervised image regression for heterogeneous change detection," *IEEE Transactions on Geoscience and Remote Sensing*, vol. 57, no. 12, pp. 9960–9975, 2019. 2, 8, 10, 12
[4] M. Gong, X. Niu, T. Zhan, and M. Zhang, "A coupling translation network for change detection in heterogeneous images," *International Journal of Remote Sensing*, vol. 40, no. 9, pp. 3647–3672, 2019. 2
[5] X. Li, Z. Du, Y. Huang, and Z. Tan, "A deep translation (gan) based change detection network for optical and sar remote sensing images," *ISPRS Journal of Photogrammetry and Remote Sensing*, vol. 179, pp. 14–34, 2021. 2
[6] X. Niu, M. Gong, T. Zhan, and Y. Yang, "A conditional adversarial network for change detection in heterogeneous images," *IEEE Geoscience and Remote Sensing Letters*, vol. 16, no. 1, pp. 45–49, 2019. 2
[7] L. T. Luppino, M. Kampffmeyer, F. M. Bianchi, G. Moser, S. B. Serpico, R. Jenssen, and S. N. Anfinsen, "Deep image translation with an affinity-based change prior for unsupervised multimodal change detection," *IEEE Transactions on Geoscience and Remote Sensing*, vol. 60, pp. 1–22, 2022. 2, 12
[8] Z. Liu, Z. Zhang, Q. Pan, and L. Ning, "Unsupervised change detection from heterogeneous data based on image translation," *IEEE Transactions on Geoscience and Remote Sensing*, pp. 1–13, 2021. 2

[9] X. Jiang, G. Li, Y. Liu, X.-P. Zhang, and Y. He, "Change detection in heterogeneous optical and sar remote sensing images via deep homogeneous feature fusion," *IEEE Journal of Selected Topics in Applied Earth Observations and Remote Sensing*, vol. 13, pp. 1551–1566, 2020. 2

[10] L. Stankovic, D. P. Mandic, M. Dakovic, I. Kisil, E. Sejdic, and A. G. Constantinides, "Understanding the basis of graph signal processing via an intuitive example-driven approach [lecture notes]," *IEEE Signal Processing Magazine*, vol. 36, no. 6, pp. 133–145, 2019. 3, 7

[11] D. I. Shuman, S. K. Narang, P. Frossard, A. Ortega, and P. Vandergheynst, "The emerging field of signal processing on graphs: Extending high-dimensional data analysis to networks and other irregular domains," *IEEE Signal Processing Magazine*, vol. 30, no. 3, pp. 83–98, 2013. 3

[12] A. Ortega, P. Frossard, J. Kovacevic, J. M. F. Moura, and P. Vandergheynst, "Graph signal processing: Overview, challenges, and applications," *Proceedings of the IEEE*, vol. 106, no. 5, pp. 808–828, 2018. 3

[13] L. Stanković, M. Daković, and E. Sejdić, "Introduction to graph signal processing," in *Vertex-Frequency Analysis of Graph Signals*. Springer, 2019, pp. 3–108. 3

[14] L. Stanković, D. Mandic, M. Daković, M. Brajović, B. Scalzo, S. Li, A. G. Constantinides *et al.*, "Data analytics on graphs part ii: Signals on graphs," *Foundations and Trends® in Machine Learning*, vol. 13, no. 2-3, 2020. 3, 7

[15] A. Sandryhaila and J. M. F. Moura, "Discrete signal processing on graphs: Frequency analysis," *IEEE Transactions on Signal Processing*, vol. 62, no. 12, pp. 3042–3054, 2014. 3

[16] P. Milanfar, "Symmetrizing smoothing filters," *SIAM Journal on Imaging Sciences*, vol. 6, no. 1, pp. 263–284, 2013. 4

[17] S. H. Chan, T. Zickler, and Y. M. Lu, "Understanding symmetric smoothing filters: A gaussian mixture model perspective," *IEEE Transactions on Image Processing*, vol. 26, no. 11, pp. 5107–5121, 2017. 4

[18] R. Sinkhorn and P. Knopp, "Concerning nonnegative matrices and doubly stochastic matrices," *Pacific Journal of Mathematics*, vol. 21, no. 2, pp. 343–348, 1967. 4

[19] X. Du, F. Nie, W. Wang, Y. Yang, and X. Zhou, "Exploiting combination effect for unsupervised feature selection by ¡inline-formula¿ ¡tex-math notation="latex"¿$\ell_{2,0}$ ¡/tex-math¿¡/inline-formula¿ norm," *IEEE Transactions on Neural Networks and Learning Systems*, vol. 30, no. 1, pp. 201–214, 2019. 6

[20] F. Nie, H. Huang, X. Cai, and C. H. Q. Ding, "Efficient and robust feature selection via joint $\ell_{2,1}$-norms minimization," in *NIPS*, 2010, pp. 1813–1821. [Online]. Available: http://papers.nips.cc/paper/3988-efficient-and-robust-feature-selection-via-joint-l21-norms-minimization 6

[21] Q. Wang, Q. Gao, X. Gao, and F. Nie, "$\ell_{2,p}$ -norm based pca for image recognition," *IEEE Transactions on Image Processing*, vol. 27, no. 3, pp. 1336–1346, 2018. 6

[22] Y. Chen and A. O. Hero, "Recursive $\ell_{1,\infty}$ group lasso," *IEEE Transactions on Signal Processing*, vol. 60, no. 8, pp. 3978–3987, 2012. 6

[23] D. Jin, J. Chen, C. Richard, and J. Chen, "Online proximal learning over jointly sparse multitask networks with $\ell_{\infty,1}$ regularization," *IEEE Transactions on Signal Processing*, vol. 68, pp. 6319–6335, 2020. 6

[24] J. Yang, W. Yin, Y. Zhang, and Y. Wang, "A fast algorithm for edge-preserving variational multichannel image restoration," *SIAM Journal on Imaging Sciences*, vol. 2, no. 2, pp. 569–592, 2009. 6

[25] N. Otsu, "A threshold selection method from gray-level histograms," *IEEE transactions on systems, man, and cybernetics*, vol. 9, no. 1, pp. 62–66, 1979. 7

[26] J. A. Hartigan and M. A. Wong, "Algorithm as 136: A k-means clustering algorithm," *Journal of the royal statistical society. series c (applied statistics)*, vol. 28, no. 1, pp. 100–108, 1979. 7

[27] J. C. Bezdek, R. Ehrlich, and W. Full, "Fcm: The fuzzy c-means clustering algorithm," *Computers & geosciences*, vol. 10, no. 2-3, pp. 191–203, 1984. 7

[28] Y. Sun, L. Lei, D. Guan, M. Li, and G. Kuang, "Sparse-constrained adaptive structure consistency-based unsupervised image regression for heterogeneous remote sensing change detection," *IEEE Transactions on Geoscience and Remote Sensing*, pp. 1–14, 2021. 7, 10, 12, 13

[29] R. Zhang, X. Li, H. Zhang, and F. Nie, "Deep fuzzy k-means with adaptive loss and entropy regularization," *IEEE Transactions on Fuzzy Systems*, vol. 28, no. 11, pp. 2814–2824, 2020. 7

[30] Y. Sun, L. Lei, X. Li, X. Tan, and G. Kuang, "Patch similarity graph matrix-based unsupervised remote sensing change detection with homogeneous and heterogeneous sensors," *IEEE Transactions on Geoscience and Remote Sensing*, vol. 59, no. 6, pp. 4841–4861, 2021. 10, 12

[31] P. Zhang, M. Gong, L. Su, J. Liu, and Z. Li, "Change detection based on deep feature representation and mapping transformation for multi-spatial-resolution remote sensing images," *ISPRS Journal of Photogrammetry and Remote Sensing*, vol. 116, pp. 24–41, 2016. 12

[32] Y. Wu, J. Li, Y. Yuan, A. K. Qin, Q.-G. Miao, and M.-G. Gong, "Commonality autoencoder: Learning common features for change detection from heterogeneous images," *IEEE Transactions on Neural Networks and Learning Systems*, pp. 1–14, 2021. 12

[33] L. Lei, Y. Sun, and G. Kuang, "Adaptive local structure consistency-based heterogeneous remote sensing change detection," *IEEE Geoscience and Remote Sensing Letters*, pp. 1–5, 2021. 12

[34] R. Touati, M. Mignotte, and M. Dahmane, "Change detection in heterogeneous remote sensing images based on an imaging modality-invariant mds representation," in *2018 25th IEEE International Conference on Image Processing (ICIP)*, 2018, pp. 3998–4002. 12

[35] R. Touati, M. Mignotte, and M. Dahmane, "Anomaly feature learning for unsupervised change detection in heterogeneous images: A deep sparse residual model," *IEEE Journal of Selected Topics in Applied Earth Observations and Remote Sensing*, vol. 13, pp. 588–600, 2020. 12

[36] R. Touati, M. Mignotte, and M. Dahmane, "A reliable mixed-norm-based multiresolution change detector in heterogeneous remote sensing images," *IEEE Journal of Selected Topics in Applied Earth Observations and Remote Sensing*, vol. 12, no. 9, pp. 3588–3601, 2019. 12

[37] Y. Zhou, H. Liu, D. Li, H. Cao, J. Yang, and Z. Li, "Cross-sensor image change detection based on deep canonically correlated autoencoders," in *International Conference on Artificial Intelligence for Communications and Networks*. Springer, 2019, pp. 251–257. 12

[38] J. Yang, Y. Zhou, Y. Cao, and L. Feng, "Heterogeneous image change detection using deep canonical correlation analysis," in *2018 24th International Conference on Pattern Recognition (ICPR)*, 2018, pp. 2917–2922. 12

[39] M. Volpi, G. Camps-Valls, and D. Tuia, "Spectral alignment of multi-temporal cross-sensor images with automated kernel canonical correlation analysis," *ISPRS Journal of Photogrammetry and Remote Sensing*, vol. 107, pp. 50–63, 2015, multitemporal remote sensing data analysis. 12

[40] Y. Sun, L. Lei, X. Tan, D. Guan, J. Wu, and G. Kuang, "Structured graph based image regression for unsupervised multimodal change detection," *ISPRS Journal of Photogrammetry and Remote Sensing*, vol. 185, pp. 16–31, 2022. [Online]. Available: https://www.sciencedirect.com/science/article/pii/S0924271622000089 12

[41] R. Touati and M. Mignotte, "An energy-based model encoding nonlocal pairwise pixel interactions for multisensor change detection," *IEEE Transactions on Geoscience and Remote Sensing*, vol. 56, no. 2, pp. 1046–1058, 2018. 12

[42] Z. Liu, G. Mercier, J. Dezert, and Q. Pan, "Change detection in heterogeneous remote sensing images based on multidimensional evidential reasoning," *IEEE Geoscience and Remote Sensing Letters*, vol. 11, no. 1, pp. 168–172, 2014. 12

[43] M. Yang, L. Jiao, F. Liu, B. Hou, S. Yang, and M. Jian, "Dpfl-nets: Deep pyramid feature learning networks for multiscale change detection," *IEEE Transactions on Neural Networks and Learning Systems*, pp. 1–15, 2021. 12

[44] T. Zhan, M. Gong, X. Jiang, and S. Li, "Log-based transformation feature learning for change detection in heterogeneous images," *IEEE Geoscience and Remote Sensing Letters*, vol. 15, no. 9, pp. 1352–1356, 2018. 12

[45] Y. Chen and L. Bruzzone, "Self-supervised change detection in multiview remote sensing images," *IEEE Transactions on Geoscience and Remote Sensing*, vol. 60, pp. 1–12, 2022. 12

[46] J. Liu, M. Gong, K. Qin, and P. Zhang, "A deep convolutional coupling network for change detection based on heterogeneous optical and radar images," *IEEE Transactions on Neural Networks and Learning Systems*, vol. 29, no. 3, pp. 545–559, 2018. 12

[47] W. Zhao, Z. Wang, M. Gong, and J. Liu, "Discriminative feature learning for unsupervised change detection in heterogeneous images based on a coupled neural network," *IEEE Transactions on Geoscience and Remote Sensing*, vol. 55, no. 12, pp. 7066–7080, 2017. 12

[48] R. Touati, M. Mignotte, and M. Dahmane, "Multimodal change detection in remote sensing images using an unsupervised pixel pairwise-based markov random field model," *IEEE Transactions on Image Processing*, vol. 29, pp. 757–767, 2020. 12

[49] M. Mignotte, "A fractal projection and markovian segmentation-based approach for multimodal change detection," *IEEE Transactions on

*Geoscience and Remote Sensing*, vol. 58, no. 11, pp. 8046–8058, 2020. 12

[50] R. Touati, "Détection de changement en imagerie satellitaire multimodale," Ph.D. dissertation, Université de Montréal, 2019. 12

[51] Y. Sun, L. Lei, X. Li, H. Sun, and G. Kuang, "Nonlocal patch similarity based heterogeneous remote sensing change detection," *Pattern Recognition*, vol. 109, p. 107598, 2021. 12

[52] Y. Sun, L. Lei, D. Guan, and G. Kuang, "Iterative robust graph for unsupervised change detection of heterogeneous remote sensing images," *IEEE Transactions on Image Processing*, vol. 30, pp. 6277–6291, 2021. 12